\documentclass{article}
\usepackage{amssymb}

\usepackage[final]{corl_2025} 
\usepackage{times} 
\usepackage{amsmath} 
\usepackage{amssymb}  
\usepackage{multirow}
\usepackage{booktabs}
\usepackage{graphicx}
\usepackage{subcaption}   
\usepackage[table,dvipsnames]{xcolor}
\usepackage[normalem]{ulem}
\usepackage{booktabs}
\usepackage{colortbl}
\usepackage{tabularray}
\usepackage{algorithm}
\usepackage{algorithmic}
\usepackage{wrapfig} 
\usepackage{float}
\usepackage{lipsum}
\usepackage{enumitem}
\usepackage{etoc} 
\usepackage{pifont}
\usepackage{bibunits}          

\definecolor{myred}{HTML}{FB4141}
\definecolor{mygreen}{HTML}{78C841}

\definecolor{fetchcolor}{RGB}{100,13,95}

\title{FetchBot: Learning Generalizable Object Fetching in Cluttered Scenes via Zero-Shot Sim2Real}

\author{
\textbf{Weiheng Liu}{\textsuperscript{\textnormal{1,3,4,*}}} \quad
\textbf{Yuxuan Wan}{\textsuperscript{\textnormal{2,3,*}}} \quad
\textbf{Jilong Wang}{\textsuperscript{\textnormal{2,3}}} \quad
\textbf{Yuxuan Kuang}{\textsuperscript{\textnormal{2}}} \quad
\textbf{Wenbo Cui}{\textsuperscript{\textnormal{1,3,4}}} \\
\textbf{Xuesong Shi}{\textsuperscript{\textnormal{3}}}  \quad
\textbf{Haoran Li}{\textsuperscript{\textnormal{1}}} \quad
\textbf{Dongbin Zhao}{\textsuperscript{\textnormal{1}}} \quad
\textbf{Zhizheng Zhang}{\textsuperscript{\textnormal{3,4,\dag}}} \quad
\textbf{He Wang}{\textsuperscript{\textnormal{2,3,4,\dag}}} \\
\textsuperscript{1}Institute of Automation, Chinese Academy of Sciences \\
\textsuperscript{2}CFCS, School of Computer Science, Peking University \\
\textsuperscript{3}Galbot \quad
\textsuperscript{4}Beijing Academy of Artificial Intelligence
}

\begin{document}
\etocdepthtag.toc{main}
\maketitle


{
\setlength{\footskip}{5pt}       
\enlargethispage{-1.5\baselineskip}  
\setlength{\skip\footins}{2pt}
\let\thefootnote\relax       
\footnotetext{* Equal contributions; author order is arbitrary and does not reflect the level of contribution. }
\footnotetext{~\,~\,Email: \url{weihliu2002@gmail.com}, \url{yaser_wyx@163.com}.}

\footnotetext{\dag~Corresponding authors. Email: \url{zhangzz@galbot.com}, \url{hewang@pku.edu.cn}.}
}

\vspace{-2.0em}
\begin{center}
    \vspace{-0.1cm}
    \captionsetup{type=figure}
    \includegraphics[trim=0.2cm 14.2cm 1cm 0.0cm, clip,  width=1.0\linewidth]{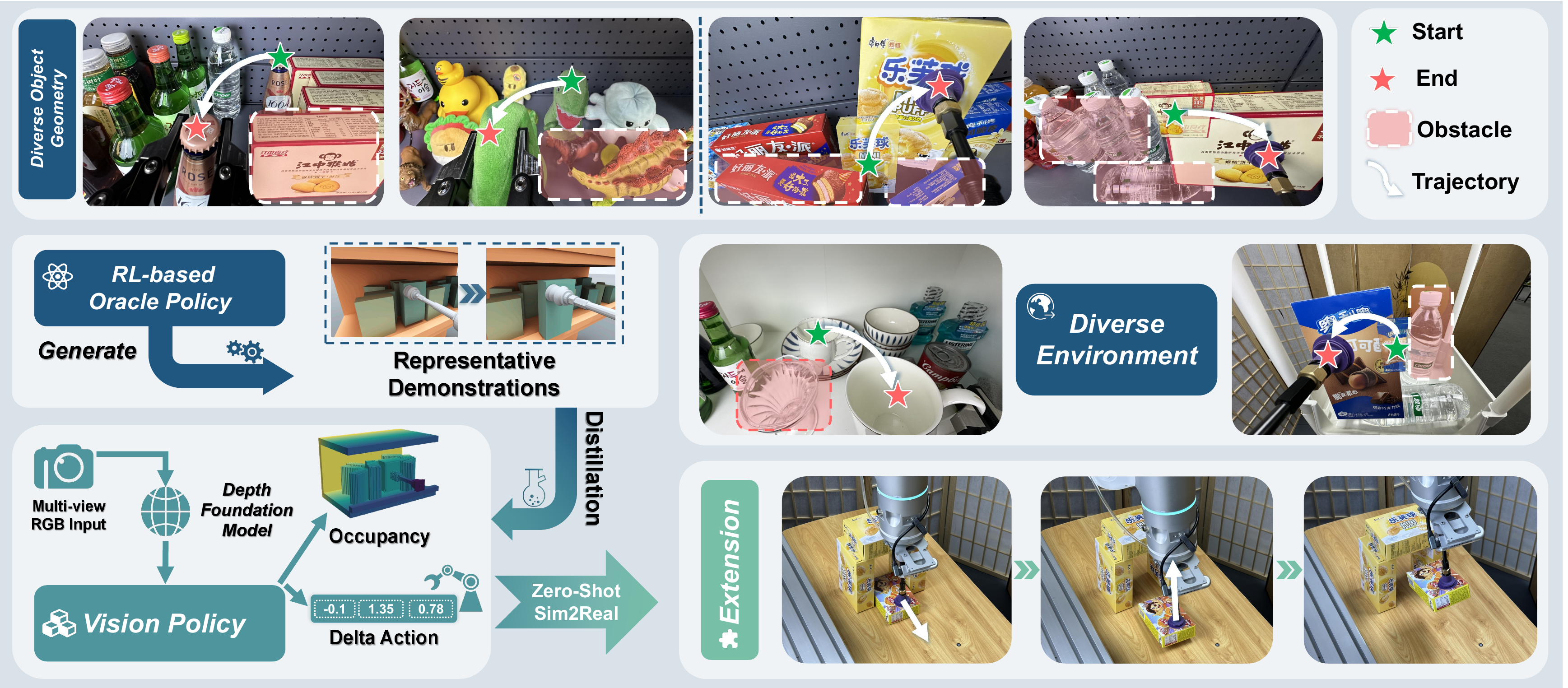}
    \captionof{figure}{FetchBot is a sim-to-real framework achieving generalizable object fetching in cluttered scenes. Its systematic design enables policy generalization and sim-to-real transferability across diverse objects, varying layouts, and multiple end-effectors. Videos are on the \href{https://pku-epic.github.io/FetchBot/}{\textcolor{fetchcolor}{\textbf{\textit{project website}}}}.
    }
    \label{fig:teaser}
    \vspace{-0.1cm}
\end{center}
\begin{bibunit}
\begin{abstract}
Generalizable object fetching in cluttered scenes remains a fundamental and application-critical challenge in embodied AI. Closely packed objects cause inevitable occlusions, making safe action generation particularly difficult. Under such partial observability, effective policies must not only generalize across diverse objects and layouts but also reason about occlusion to avoid collisions. However, collecting large-scale real-world data for this task remains prohibitively expensive, leaving this problem largely unsolved. In this paper, we introduce FetchBot, a sim-to-real framework for this challenge. We first curate a large-scale synthetic dataset featuring 1M diverse scenes and 500k representative demonstrations. Based on this dataset, FetchBot employs a depth-conditioned method for action generation, which leverages structural cues to enable robust obstacle-aware action planning. However, depth is perfect in simulation but noisy in real-world environments. To address this sim-to-real gap, FetchBot predicts depth from RGB inputs using a foundation model and integrates local occupancy prediction as a pre-training task, providing a generalizable latent representation for sim-to-real transfer. Extensive experiments in simulation and real-world environments demonstrate FetchBot’s strong zero-shot sim-to-real transfer, effective clutter handling, and adaptability to novel scenarios. In cluttered environments, it achieves an average real-world success rate of 89.95\%, significantly outperforming prior methods. Moreover, FetchBot demonstrates excellent robustness in challenging cases, such as fetching transparent, reflective, and irregular objects, highlighting its practical value. Project website: \href{https://pku-epic.github.io/FetchBot/}{\textcolor{fetchcolor}{\textbf{\textit{https://pku-epic.github.io/FetchBot/}}}}.

\end{abstract}
\keywords{Generalizable Fetching, Sim2Real, Occlusion Handling}

\section{Introduction}
\vspace{-0.2cm}
Cluttered scenes are ubiquitous—from densely packed retail displays and disorganized warehouse racks to crowded kitchen cabinets—making reliable object fetching in such environments an essential capability for embodied AI applications. 
Densely arranged objects introduce severe occlusions and rich obstacles, requiring an effective fetching policy to reason about object geometry and obstacles to minimize collisions under partial observation while generalizing across diverse categories, layouts, and materials. These demands make generalizable fetching in cluttered scenes a largely unsolved problem.


For this task, some prior works~\cite{atar2024optigrasp,yang2023dynamo,spahn2024demonstrating,bajracharya2024demonstrating,murray2024learning}  
that employ heuristics or motion planning based on partial environment observation often fall short in highly cluttered scenes, especially when no collision-free path exists.
Other works~\cite{wada2022safepicking,han2024fetchbench} investigate learning-based methods, however they struggle to handle the generalization across diverse object arrangements, categories, geometries, and materials commonly encountered in real-world scenarios due to limited amount of real data~\cite{kaplan2020scaling,hoffmann2022training,chen2024internvl}. 
In contrast, synthetic data offers a more efficient and scalable alternative. Some recent works~\cite{han2024fetchbench,wada2022safepicking} propose to augment existing fetching datasets in simulation environments.
However, the generated scenes are overly sparse, featuring widely spaced objects with minimal occlusion. Therefore, the policies fail to generalize to the densely cluttered environments encountered in the real world. Furthermore, sim-to-real gap persists. Several methods~\cite{lyu2024scissorbot,xie2023part,qin2023dexpoint,wang2025mobileh2r} leverage depth and point-cloud inputs to bridge this gap, given its focus on geometry instead of texture~\cite{zhang2023close}.
Nevertheless, practical limitations still remain: real depth sensors often produce flying pixels around object boundaries~\cite{park2023edge}, unreliable measurements on reflective and transparent surfaces, and substantial noise. These factors collectively undermine robust performance in complex real-world scenarios.

Along with generalization, fetching safety, which requires environmental impact minimization, is crucial since slight collisions may lead to hazards (illustrated in Fig.~\ref{fig:teaser}). 
Works~\cite{shridhar2023perceiver,goyal2023rvt,ke20243d,liu2024voxact} model environments using scene voxel maps, typically constructed by projecting RGB views via sensed depth and camera extrinsics. This technique directly inherits the limitations of depth sensors, which are often unreliable in practice. Critically, it also produces incomplete voxel representations in heavily occluded regions, thereby losing essential geometric details. 

To address the above challenges, we introduce FetchBot, a 
framework designed for generalizable and safe robotic fetching in occlusion-rich environments. 
To mitigate data scarcity, we develop a Unified Voxel-based Scene Generator (UniVoxGen) to synthesize millions of cluttered scenes efficiently, where objects are tightly packed and frequently occlude one another. 
Given that motion planning often fails to find collision-free trajectories in such scenes, we train a dynamics-aware oracle policy using reinforcement learning to generate representative demonstrations. 
We then leverage depth predicted by a foundation model~\cite{yang2024depth} as an intermediate representation to bridge the sim-to-real gap.
To tackle the incomplete scene understanding caused by heavy occlusion, we integrate occupancy prediction~\cite{wang2024panoocc,wei2023surroundocc,cao2022monoscene} as an auxiliary task to overcome perception limitations arising from heavy occlusion. This task encourages the model to infer occluded regions through spatial reasoning, leading to more complete scene understanding.
Through extensive simulation and real-world experiments, our method outperforms existing approaches by a significant margin, achieving an average 89.95\% success rate in real-world scenarios. This robustness also extends to challenging objects, including translucent, reflective, and irregular items.

In summary, we make the following contributions: 1) Generate a large-scale dataset comprising 1M diverse cluttered scenes using UniVoxGen and 500k demonstrations using a dynamics-aware oracle policy. This dataset serves as a foundation for developing generalizable fetching skills.
2) Introduce a depth-conditioned action generation policy. This policy leverages depth predictions from a foundation model to bridge the sim-to-real gap. Additionally, it integrates occupancy prediction as a pre-training task, enabling the model to infer information about occlusions and achieve comprehensive scene understanding.
3) Validate our method through extensive simulation and real-world experiments, demonstrating significant improvements in handling cluttered scenes, achieving zero-shot sim-to-real transfer, and generalizing across diverse scenarios.
\section{Related Works}


\textbf{Robotic Fetching from Cluttered Scenes.}
Robotic fetching is a long-standing and fundamental challenge within robotic manipulation, attracting extensive research interest over the years \cite{newbury2023deep}. A particularly demanding aspect of this challenge involves fetching objects from cluttered environments \cite{correll2016analysis,eppner2016lessons,mahler2017learning,mahler2019learning,yu2016summary,li2023sim,murray2024learning,yang2023dynamo,atar2024optigrasp,han2024fetchbench,zhang2024gamma,wang2024quadwbg}. 
While numerous studies \cite{murray2024learning,yang2023dynamo,atar2024optigrasp} have focused on identifying initial picking or grasping points, the subsequent, critical retrieval stage is often under-emphasized. This phase necessitates careful maneuvering to minimize disturbance to surrounding objects, as even slight collisions can destabilize the scene. 
Recently, benchmarks like FetchBench~\cite{han2024fetchbench} have begun to specifically target the complexities of object fetching. However, generating simulated environments that fully capture the high density and unpredictable nature of real-world clutter remains difficult, potentially limiting the direct applicability of models trained in such settings. Addressing this gap, our work employs a voxel-based method to generate simulated cluttered scenes that more realistically capture the complexity of real-world environments.

\textbf{Sim2Real Transfer for 3D Visuomotor Policies.} Currently, there are many 3D-based imitation learning policies \cite{shridhar2023perceiver,goyal2023rvt,gervet2023act3d,ze20243d,ke20243d,xian2023chaineddiffuser,liu2024voxact} that utilize 3D observation data to mimic expert actions from demonstrations. However, these methods predominantly rely on real-world data to perform real-robot tasks, failing to fully leverage the potential of simulators. As a result, sim-to-real transfer for 3D visuomotor policies remains an under-explored topic. Most previous works~\cite{lyu2024scissorbot,xie2023part,qin2023dexpoint,wang2025mobileh2r,lin2024prompting} have employed point clouds as representations to achieve sim-to-real, but they still struggle to bridge the sim-to-real gap due to noise and inaccuracies, particularly at object edges and reflective surfaces in real-world point clouds captured by depth sensors. To further advance the field of sim-to-real research, inspired by approaches in autonomous driving \cite{wang2024panoocc,cao2022monoscene,wei2023surroundocc} and computer vision foundation model~\cite{yang2024depth,godard2019digging,hu2024depthcrafter,shao2024learning, irshad2024nerf}, we propose utilizing predicted depth as an intermediate representation, leveraging its strong generalization ability to mitigate the sim-to-real gap. Furthermore, we employ a unified 3D representation to ensure consistent multi-view image fusion, thereby achieving complete perception and further minimizing the sim-to-real gap.

\section{FetchBot}
\subsection{Overview}
To enable generalizable and safe fetching skills, we introduce FetchBot, a framework that learns a generalizable fetching policy through zero-shot sim-to-real.
We begin by efficiently generating realistic, densely cluttered scenes using a voxel-based method (\S~\ref{scene_generator}), and collecting demonstrations from a dynamics-aware oracle policy (\S~\ref{oracle_policy}). Next, we leverage the depth predicted by a foundation model as the intermediate representation to bridge the sim-to-real gap (\S~\ref{foundation_model}). Finally, we introduce an occupancy-based voxel representation to address the challenge of incomplete perception caused by heavy occlusion (\S\ref{Occlusion_Completion}). 

\begin{figure*}[tb]
    \centering
    \includegraphics[trim=0cm 22.5cm 8.5cm 0.0cm, clip,  width=1.0\linewidth]{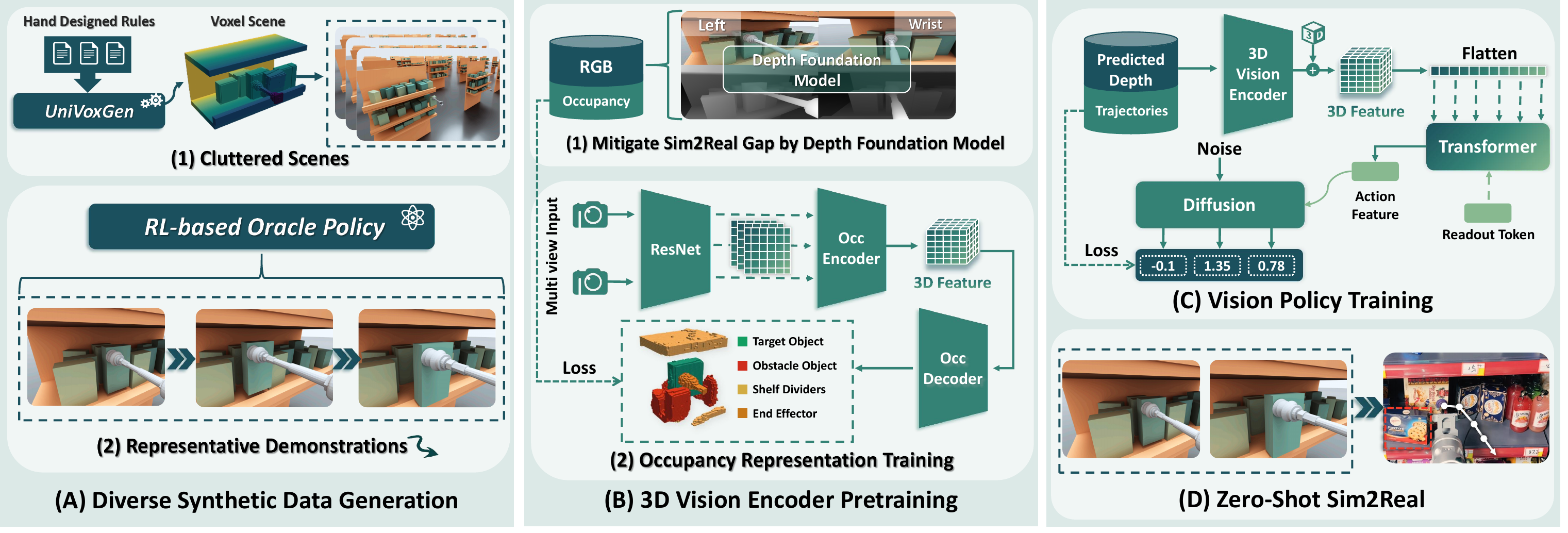}
    \caption{\textbf{FetchBot Pipeline.} In the (A) data generation stage, we use UniVoxGen to generate a diverse set of cluttered scenes and employ an RL-based Oracle policy to collect representative demonstrations. In the (B) 3D vision encoder pretraining stage, we first use the foundation model's predicted depth as an intermediate representation to mitigate the sim-to-real gap, then introduce an occupancy prediction task to learn a complete scene representation that can infer occluded regions. In the (C) vision policy training stage, we distill these expert demonstrations into a vision-based policy through imitation learning, which can achieve (D) zero-shot sim-to-real.}
    \label{fig_2}
    \vspace{-0.5cm}
\end{figure*}

\subsection{Voxel-based Cluttered Scene Generator}
\label{scene_generator}

Accurate scene generation mandates collision-free object placement. However, existing methods~\cite {wada2022safepicking, li2024broadcasting, xu2023joint, qian2024thinkgrasp,yang2024ground4act,wang2024learning} often rely on computationally intensive simulator-based collision checks or simulated object dropping. These techniques are prone to generating unrealistic configurations (see Appendix~\ref{appendix_a} for details) and significantly hinder both generation efficiency and scene validity, particularly in densely cluttered and occluded environments. 
\begin{wrapfigure}{r}{0.325\textwidth}
  \centering
  \includegraphics[trim=0cm 31.5cm 46.8cm 0cm, clip, width=\linewidth]{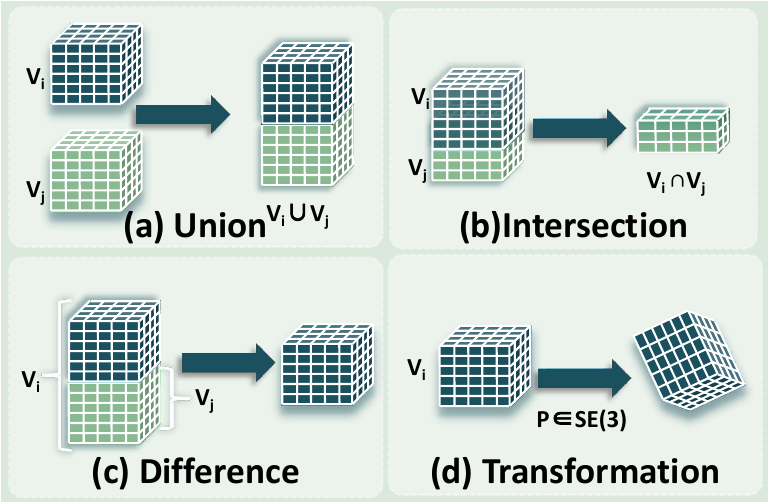}
  \caption{A set of fundamental voxel operations in UniVoxGen.}
  \label{fig_6}
 \vspace{-5pt} 
  
\end{wrapfigure}
In contrast, our approach, UniVoxGen, operates within a unified voxel space established by initially voxelizing each object. As depicted in Fig.~\ref{fig_6}, we leverage fundamental voxel operations: \emph{Union} (adding objects), \emph{Intersection} (detecting collisions), \emph{Difference} (removing objects), and \emph{Transformation} (adjusting object poses). These operations are computationally lightweight and execute rapidly, facilitating the efficient generation of realistic scenes. Leveraging this efficiency, we employed UniVoxGen to create a large-scale dataset of 1 million cluttered scenes, which serve as training data for our oracle policy. Furthermore, the generated voxel scenes provide dense ground truth for training the occupancy prediction task.

\subsection{Dynamics-Aware Oracle Policy}
\label{oracle_policy}

To generate demonstrations with minimal disturbance, we train an oracle policy using reinforcement learning. Through extensive interaction, this policy learns implicit environment dynamics, enabling it to understand the impact of potential actions and select optimal, low-impact maneuvers. A core component of this policy is a hierarchical scene encoder designed to capture both local object details and global scene context, allowing for effective representation of complex clutter.




\textbf{Observations Encoding.}
The observation $\boldsymbol{o}_t$ of oracle policy $\boldsymbol{\pi}$ combines proprioception $\boldsymbol{p}_t$, the previous action $\boldsymbol{a}_{t-1}$, and a scene representation $\boldsymbol{z}_t$. 
\begin{wrapfigure}{r}{0.50\textwidth}
    \vspace{-0.3cm}
  \centering
  \includegraphics[trim=0cm 33.8cm 41.5cm 0.0cm, clip, width=0.50\textwidth]{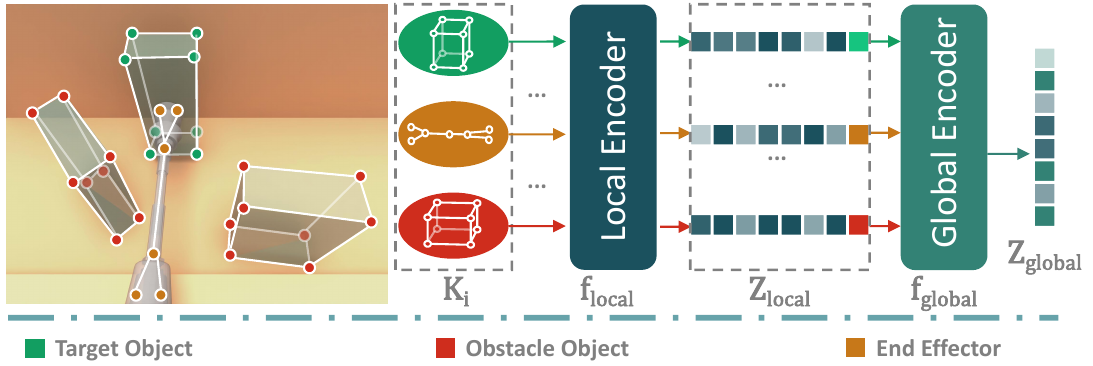}
    \vspace{-0.5cm}
  \caption{\textbf{Scene Encoder Network.} The network employs a hierarchical architecture, where a local network extracts object features and a global network captures scene-level structure and context.} 
  \vspace{-0.5cm}
  \label{fig_3}
\end{wrapfigure}
Inspired by PointNet++~\cite{qi2017pointnet++}, we adopt a hierarchical network $\boldsymbol{f}_{scene}$. As depicted in Fig.~\ref{fig_3}, 
consider a scenario $\boldsymbol{S}$ comprising a set of objects, and each object is represented by a set of keypoints $\boldsymbol{K}_i = \{\boldsymbol{k}_{i,1}, \boldsymbol{k}_{i,2}, \ldots\, \boldsymbol{k}_{i,n}\}$. 
Our hierarchical process first extracts local features $\boldsymbol{z}_{local}^{(i)}$ for each object independently using a network $\boldsymbol{f}_{local}$: $\boldsymbol{z}_{local}^{(i)} = \boldsymbol{f}_{local}(K_i)$. 
Subsequently, these local features are aggregated by a global network $\boldsymbol{f}_{global}$ to produce the final scene representation $\boldsymbol{z} = \boldsymbol{z}_{global}$: $\boldsymbol{z}_{global} = \boldsymbol{f}_{global}(\{\boldsymbol{z}_{local}^{(1)}, \ldots, \boldsymbol{z}_{local}^{(N)}\})$. 
This global feature $\boldsymbol{z}$ serves as our comprehensive scene representation. The networks $\boldsymbol{f}_{local}$ and $\boldsymbol{f}_{global}$ each comprise a two-layer MLP followed by max pooling. This design enables permutation invariance, therefore enhancing robustness. This hierarchical architecture could preserve detailed object information alongside global scene understanding.

\textbf{Reward.}
Our reward function combines task completion, behavioral constraints, and an environment change penalty based on per-step obstacle movement $m_{step}$ (translation and rotation): $r = \lambda_{task} r_{task} + \lambda_{cons} r_{cons} + \lambda_{env} r_{env}$.
Task reward is given for successful fetching, but only if the total obstacle change per episode $m_{episode}$ is less than $\sigma_m$, a success threshold. We employ a curriculum for $\sigma_m$, starting high to facilitate initial task completion and gradually decreasing it to enforce precision and minimal environmental impact later in training. Detailed reward design is provided in Appendix~\ref{appendix_b}.

\subsection{Vision-based Imitation Learning}

\textbf{Intermediate Depth Representation for Sim-to-Real.}
\label{foundation_model}
Depth maps offer superior spatial information compared to RGB images, which is crucial for robotic fetching.
Nonetheless, direct sim-to-real policy transfer with depth is difficult due to depth sensor limitations like edge inaccuracies, poor transparency handling, and substantial noise.
To address this, we utilize a sim-to-real-friendly intermediate depth representation derived from a depth foundation model. We use DepthAnything~\cite{yang2024depth} to map both simulation RGB images $P_s$ and real-world RGB images $P_r$ to a shared depth space $D_f$. 
This intermediate representation ensures the policy always receives input within the same input space $D_f$. Consequently, this design allows FetchBot to maintain consistent action predictions across domains and significantly improve its sim-to-real capability.

\textbf{Completion of Occluded Regions.}
\label{Occlusion_Completion}
Planning safe trajectories in cluttered scenes is challenging due to heavy occlusion hindering complete perception.
As depicted in Fig.~\ref{fig_2} (B), we address this by introducing semantic occupancy prediction as an auxiliary task.
After obtaining multi-view predicted depth maps from the DepthAnything, we extract feature maps $I = \{i_i\}_{i=1}^N$ using a ResNet-50 backbone. These features are subsequently aggregated into a 3D latent space using a multi-view feature fusion mechanism. Specifically, within the 3D vision encoder, we define a set of learnable local 3D-grid queries $Q \in \mathbb{R}^{C \times H \times W \times Z}$, where $H$, $W$, and $Z$ denote the query grid dimensions. For each 3D query point $p$, we project it into the 2D feature maps $\mathcal{F}^{2D}$ according to known camera parameters, utilizing only information from the valid views $V_{hit}$. A deformable cross-attention (DCA) mechanism then aggregates local 2D features around the projected points:
\vspace{-0.5pt}
$$\text{DCA}(q_p, \mathcal{F}^{2D}) = \frac{1}{|V_{hit}|} \sum_{t \in V_{hit}} \text{DA}(q_p, \mathcal{P}(p,t), \mathcal{F}_t^{2D}),$$
where $q_p$ is the feature associated with 3D point $p=(x,y,z)$, $\mathcal{P}(p,t)$ is the projection function, $\mathcal{F}_t^{2D}$ is the 2D feature map from hitted view $t$, and $\text{DA}$ denotes deformable attention~\cite{zhu2020deformable}. The aggregated 3D features are further refined using 3D convolutions (see Appendix~\ref{appendix_c} for more details).

Within a local 3D volume around the end-effector, referred to as the region of interest (ROI), we predict semantic occupancy, distinguishing object classes (target, obstacle, robot) instead of simply identifying occupied or free spaces. This semantic representation is crucial for identifying the target object in cluttered scenes. Local prediction within ROI improves both computational efficiency and generalization by exposing the network to diverse partial observations (irregular shape due to cropping). Additionally, the network uses multi-view predicted depth maps as input to predict occupancy in unobserved regions, and the model is supervised with complete scene occupancy (generated by UniVoxGen). This auxiliary task enables the model to infer the full structure from limited cues, enhancing robustness to occlusions in cluttered scenes.

As illustrated in Fig.~\ref{fig_2} (C), building upon the 3D semantic occupancy prediction, our policy translates this scene representation \( \mathcal{F}^{3D} \) into executable actions through a transformer-based architecture~\cite{team2024octo,kim24openvla,wang2024rise}. Specifically, we augment the 3D features with learnable position embeddings \( P^{3D} \in \mathbb{R}^{C \times H \times W \times Z} \) before flattening them into feature vectors \( V^{3D} \in \mathbb{R}^{C} \). A learnable readout token queries and aggregates these features to produce action features \( \mathcal{F}_A \). Finally, a lightweight diffusion head~\cite{chi2023diffusion} denoises Gaussian noise $a^K$ into the predicted action \( a^0 \), conditioned on \( \mathcal{F}_A \).



\textbf{Two-Stage Policy Learning Framework.} 
We begin by pretraining the 3D vision encoder on UniVoxGen’s complete occupancy data, optimizing cross-entropy and scene-class affinity losses~\cite{cao2022monoscene} to establish robust scene understanding.
In the second stage, we freeze the pre-trained vision encoder. Concurrently, the policy head (comprising transformer and diffusion components) is trained via imitation learning. This learning process minimizes the mean squared error (MSE) between the predicted noise and the actual noise applied to the oracle actions.

\section{Experiments}

\begin{figure*}[tb]
    \centering
    \includegraphics[trim=0cm 15.5cm 5.5cm 0.0cm, clip,  width=0.95\linewidth]{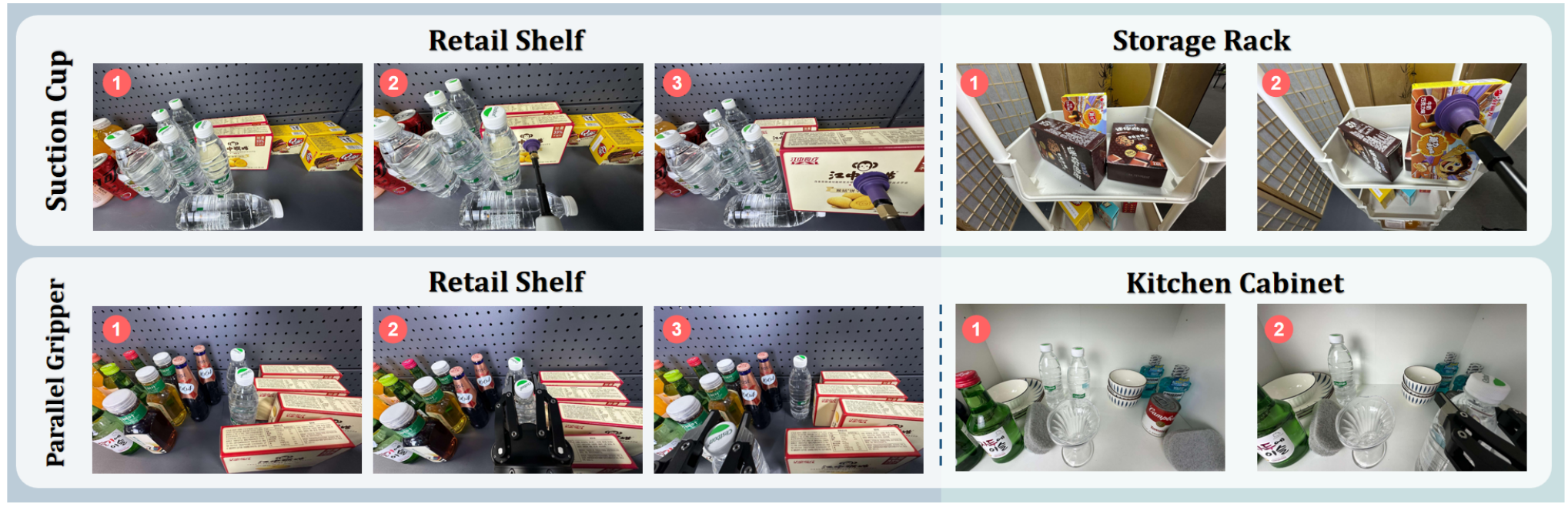}
    \caption{\textbf{Qualitative Results in the real world.} We evaluate our system in diverse and cluttered scenes containing a variety of objects and environments. The upper row shows a suction cup, and the bottom row shows the parallel gripper.}
    \label{fig_5}
\end{figure*}

\begin{table}
\centering
\resizebox{0.98\linewidth}{!}{
    \begin{tabular}{l|ccc|ccc} 
    \toprule
    \textbf{End-effector}    & \multicolumn{3}{c|}{\textbf{Suction Cup}}                                 & \multicolumn{3}{c}{\textbf{Parallel Gripper}}                              \\ 
    \midrule
    \textbf{Method}  & \textbf{Success Rate (\%) $\uparrow$} & \textbf{Translation (cm) $\downarrow$} & \textbf{Rotation (rad) $\downarrow$} & \textbf{Success Rate (\%) $\uparrow$} & \textbf{Translation (cm) $\downarrow$} & \textbf{Rotation (rad) $\downarrow$}  \\ 
    \midrule
    Heuristc         &         54.98             &        6.71             &        0.91           &       63.43                &     0.41                 &     0.71              \\
    CuRobo           &         68.26             &        3.96             &        0.63           &        73.05               &     0.24                 &     0.49               \\
    AIT*             &          62.52            &        5.86             &        0.75           &         67.87              &       0.35               &      0.63              \\
    \textbf{Oracle}  &         \textbf{85.60}    &         \textbf{1.47}   &       \textbf{0.29}   &      \textbf{94.11}         &        \textbf{0.13} &     \textbf{0.02}               \\ 
    \midrule
    RGB (DP)         &          70.35            &          4.97            &      0.55             &        75.85               &    0.24                  &    0.13                \\
    Point Cloud (DP3) &         72.37            &          4.49            &      0.76             &        80.45              &     0.21                 &      0.16              \\
    Raw Depth        &          71.44            &          4.49            &       0.56            &        71.44               &     0.17                 &     0.09               \\
    Predicted Depth  &          61.21            &          6.54            &       0.76            &        68.37               &     0.38                 &     0.12               \\
    RGB-D Voxel      &          71.38            &          4.65            &       0.62            &         79.73              &      0.22                &      0.10              \\
   \textbf{Occupancy (Ours)}  &  \textbf{81.46}  &    \textbf{2.78}       &     \textbf{0.36}     &         \textbf{91.02}          &      \textbf{0.13}            &    \textbf{ 0.06}               \\
    \bottomrule
    \end{tabular}
}
\vspace{0.2cm}
\caption{\textbf{Simulation Results.} Compares the performance of different methods across two end effectors (suction cup and parallel gripper) in terms of success rate, translation disturbance, and rotation disturbance. Our method outperforms all baselines (except for the oracle), achieving the highest success rate and the lowest disturbance in translation and rotation.}
\label{table_1}
\vspace{-0.6cm}
\end{table}
In this section, we conduct extensive experiments to validate the effectiveness of our method. Our primary evaluations focus on shelf environments, which present the most significant challenges for fetching tasks. First, we compare our method with several baselines in the Isaac Gym~\cite{isaacgym} simulator. Next, we deploy the system in a replicated retail store to confirm the sim-to-real performance. Finally, we conduct ablation studies to demonstrate the contribution of each key component to the sim2real transfer. We use two end effectors, a parallel gripper for bottle-shaped items and a suction cup for box-shaped items, to demonstrate our method's flexibility.

\subsection{Simulation Experiment}
In simulation experiments, we utilize 3000 densely constructed scenes to assess safe fetching performance under occlusion. We further investigate the efficacy of different representation methods for fetching in these cluttered environments by modifying the input modality module.


\textbf{Metrics.} Our simulation evaluations assess two key metrics: (1) environmental impact, measured by total obstacle displacement (both translational and rotational), evaluates fetching-induced disturbances, and (2) success rate, requiring target fetching with obstacle displacement kept under a 3 cm threshold, which we determined to be an acceptable tolerance during fetching.

\textbf{Baselines.} 
We compare FetchBot with three series of methods: (1) heuristic-based methods, utilizing rules to generate trajectories composed of straight-line paths. (e.g., lift-then-extract), (2) motion planning-based methods (CuRobo~\cite{sundaralingam2023curobo} and AIT*~\cite{strub2020adaptively}), which find collision-free paths for fetching via motion planners, and (3) learning-based methods, including the original RGB-based diffusion policy (DP~\cite{chi2023diffusion}), 3D point cloud-based DP3~\cite{ze20243d}, RGB-D voxel-based diffusion policy, raw depth image-based diffusion policy, and DepthAnything-based diffusion policy, all these baselines employ the same diffusion denoising process for action generation (see Appendix~\ref{appendix_d} for more details).

\begin{wrapfigure}{r}{0.45\textwidth}
\vspace{-0.4cm}
  \centering
  \includegraphics[trim=0cm 28cm 27cm 0.0cm, clip, width=0.45\textwidth]{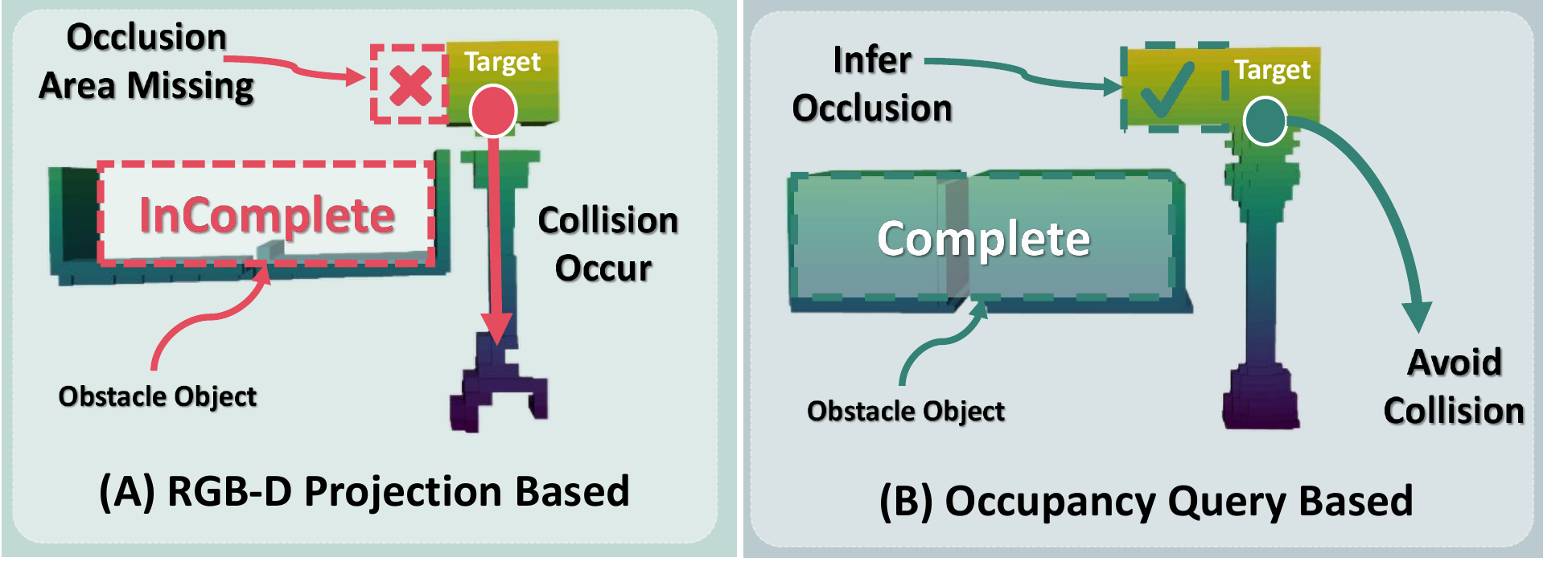}
  \caption{\textbf{Comparison in simulation.} \textbf{(A)} RGB-D voxel method misses crucial geometric details due to occlusion, leading to collision during fetching. \textbf{(B)} Our method can infer the occluded region, enabling successful collision avoidance.} 
  \label{fig_4}
  \vspace{-0.4cm}
\end{wrapfigure}

\textbf{Results and Analysis.}
As shown in Table~\ref{table_1}, heuristic-based methods exhibit the poorest performance due to their complete disregard for environmental information and reliance on predefined trajectories, resulting in frequent collisions. Motion planning-based approaches also demonstrate limited effectiveness, failing to reliably generate collision-free paths in cluttered scenes as their planners account only for static geometric configurations while neglecting the dynamic consequences of fetching operations. 
Conversely, our oracle policy learns underlying environmental dynamics through extensive interactions, enabling it to select actions that minimize environmental change.
Learning-based methods achieve substantially superior performance, though with notable constraints: RGB (DP) policy encounters obstacle avoidance difficulties stemming from absent spatial geometric information, while point cloud, depth-based, and RGB-D voxel methods, despite their enhanced scene representations, are hindered by partial observability induced by severe occlusions, ultimately diminishing success rates and environmental impact metrics. As shown in Fig.~\ref{fig_4} (A), due to occlusion, the RGB-D voxel method fails to capture the hidden part of the target, resulting in a naive extraction attempt and a subsequent collision with obstacles.
Besides struggling with heavy occlusion, the DepthAnything-based method's performance is further hindered by its scale-ambiguous depth predictions~\cite{yang2024depth}, yielding only relative depth, inadequate for true geometric representation.
By contrast, FetchBot's occupancy representation effectively integrates multi-view spatial information and uses contextual reasoning to infer geometric details of occluded areas, as depicted in Fig.~\ref{fig_4} (B). This approach leads to a more complete and structured 3D scene understanding, which facilitates object fetching in complex environments, ultimately achieving a higher success rate of 81.46\% for suction cup and 91.02\% for parallel gripper. Furthermore, FetchBot excels in the environmental impact metric, meaning it effectively considers how each action affects nearby objects and chooses actions that minimize disturbance.

\subsection{Real World Experiment}

\begin{wrapfigure}{r}{0.43\textwidth}
  \centering
    \vspace{-1.05cm}
  \includegraphics[trim=0cm 17cm 33.5cm 0.0cm, clip, width=0.4\textwidth]{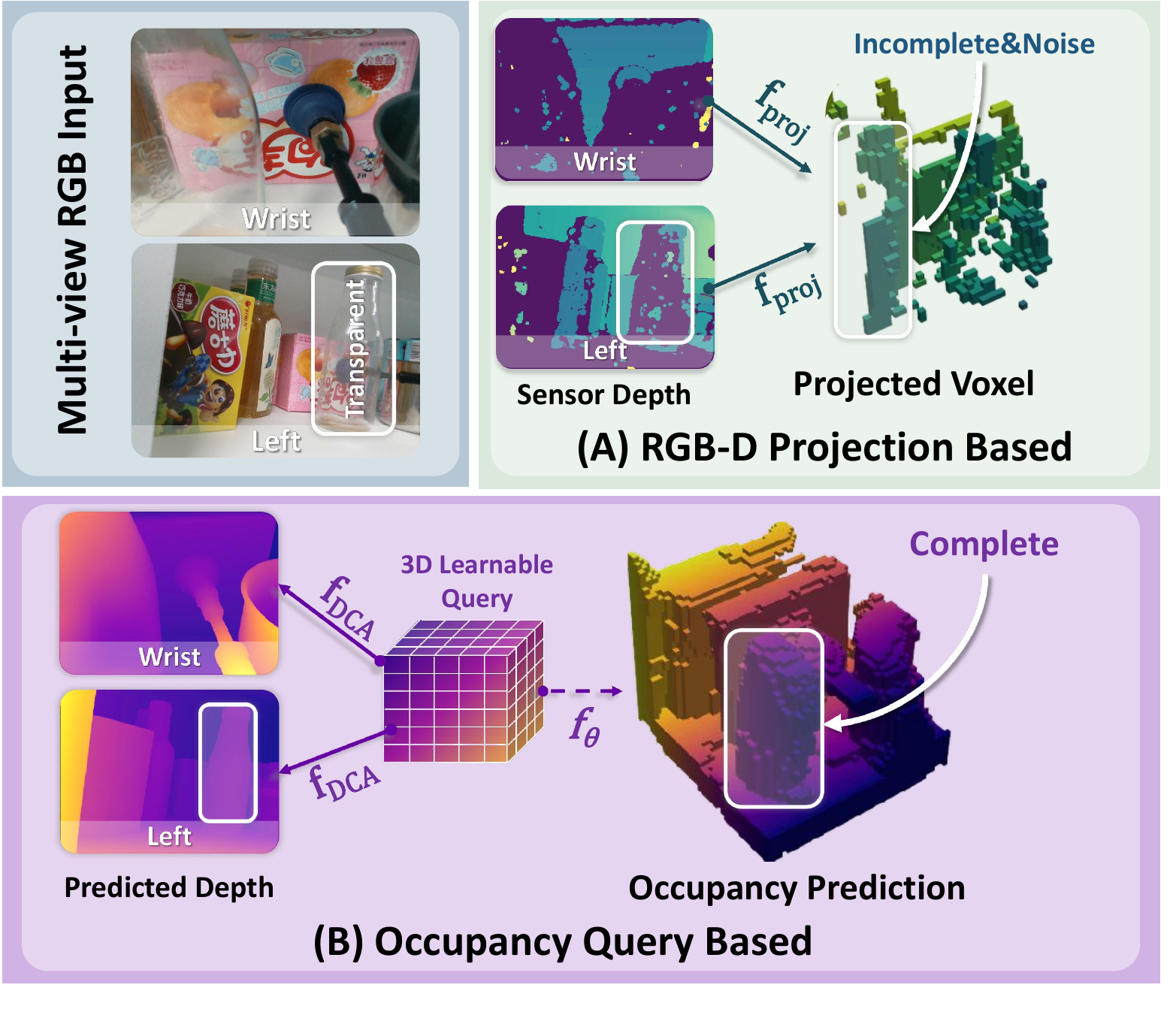}
    \vspace{-0.2cm}
  \caption{Comparison in the real world. Direct RGB-D voxelization often yields incomplete scenes in real-world settings, while our occupancy method (Occ) generates more complete representations.} 

  \label{fig_real_compare_voxel_occ}
  \vspace{-0.8cm}
\end{wrapfigure}
Extensive real-world experiments validated each method's generalization and zero-shot sim-to-real transfer across diverse scenarios (varied objects, challenging materials, environments) using 15 trials with both suction and gripper end-effectors (All scenes are shown in Appendix~\ref{appendix_e}).

\textbf{Metrics.}
Due to the unavailability of precise environmental measurements in real-world settings, the success rate serves as the only evaluation metric. 
A real-world trial is counted as successful only if the target is acquired and no surrounding objects exhibit visible displacement during the fetch.

\textbf{Baselines.} Our real-world experimental evaluation focuses exclusively on learning-based approaches, employing identical representation modalities to those used in simulation (RGB, point clouds, raw depth, DepthAnything, and RGB-D voxel).

\begin{wrapfigure}{r}{0.5\textwidth}
\vspace{-0.3cm}
  \centering
\includegraphics[trim=1.cm  28cm 26.2cm 0.6cm, clip,  width=0.95\linewidth]{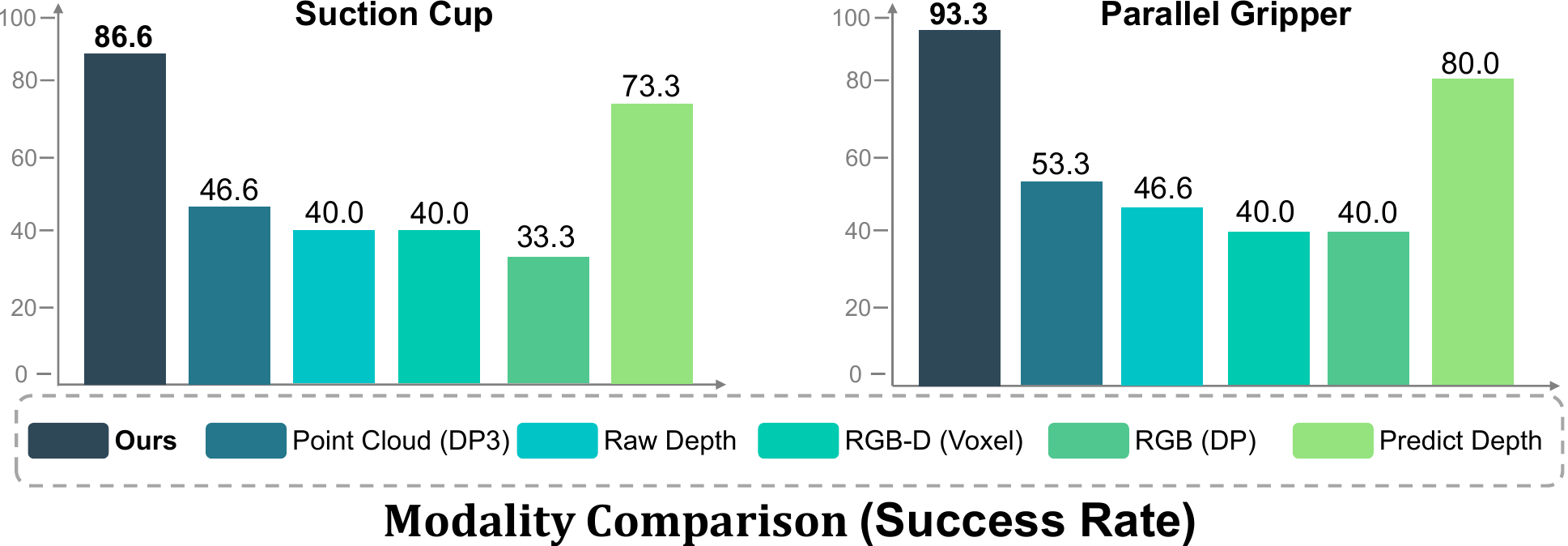}
  \caption{Modalities comparison in the real world.} 
  \label{fig_modality}
  \vspace{-0.5cm}
\end{wrapfigure}

\begin{wrapfigure}{r}{0.5\textwidth}
\vspace{-2.cm}
  \centering
  \includegraphics[trim=0cm 13.5cm 19cm 0.5cm, clip, width=0.45\textwidth]{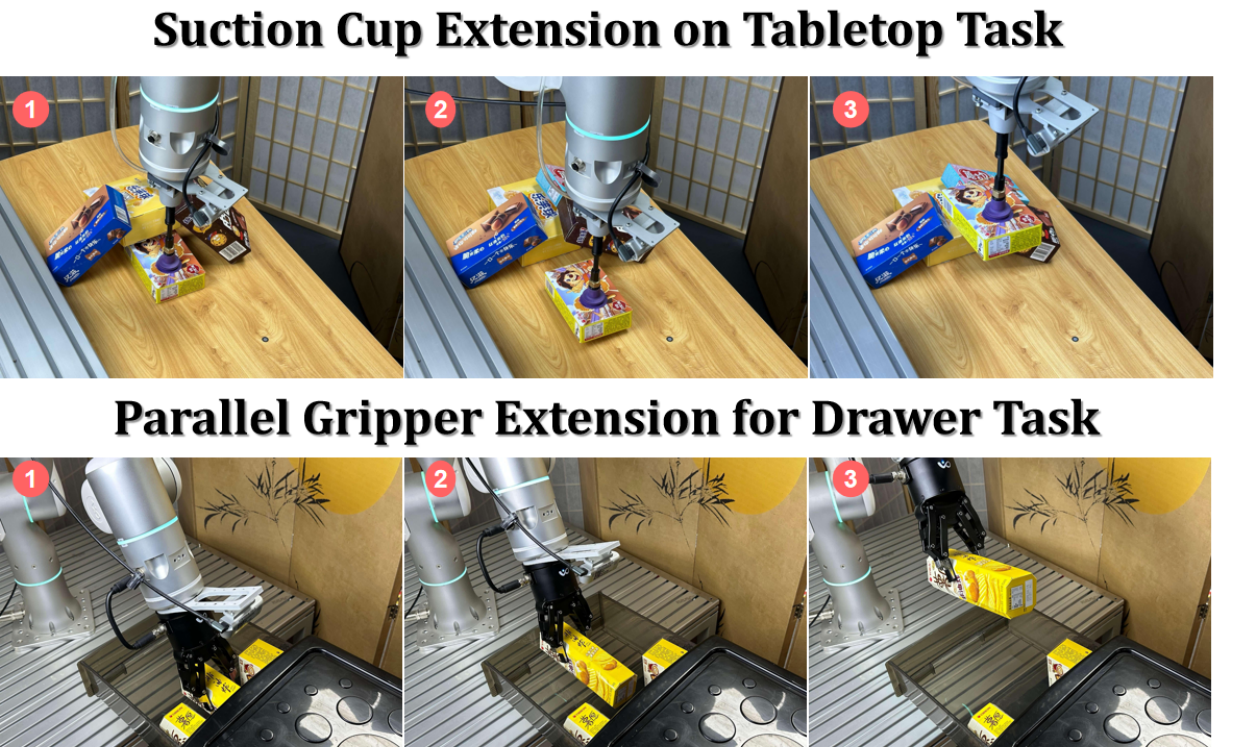}
  \caption{Real-robot extension experiments showing successful fetching from cluttered tabletop (suction) and drawer (parallel gripper) settings.} 
  \label{fig_7}
  \vspace{-0.5cm}
\end{wrapfigure}

\textbf{Results and Analysis.} 
Fig.~\ref{fig_modality} presents the zero-shot sim-to-real performance of each method in the real world. Due to differences in object textures, lighting, and other factors between simulation and reality, the RGB (DP) still underperforms despite extensive domain randomization. 
Additionally, the impact of depth noise and the inherent limitations of depth sensors (e.g., poor performance on transparent and specular materials) limit the effectiveness of point cloud and raw depth-based methods in zero-shot sim-to-real transfer, as illustrated in Fig.~\ref{fig_real_compare_voxel_occ} (A). 
While the DepthAnything-based method uses an intermediate representation to predict depth from both sim and real-world images, partially addressing the sim-to-real gap, its predicted depth is scale-ambiguous~\cite{yang2024depth}, which ultimately limits the method's overall performance. 
Moreover, limited observations due to cluttered scenes further restrict its deployment. 
In contrast, FetchBot leverages predicted depth from a foundation model to bridge the sim-to-real gap and handle challenging objects like transparent or reflective ones. It also uses an occupancy-based representation to infer occluded regions, providing complete and structured geometry for better spatial understanding, as shown in Fig.~\ref{fig_real_compare_voxel_occ} (B).
By focusing on local occupancy and semantics within ROI and extensively randomizing object placements there during training, the system generalizes effectively by exposing the policy to a wide variety of local geometries.
Its success in varied tasks, including tasks like tabletop fetching, as well as drawer extraction without collision, demonstrates broad applicability beyond shelf setting (shown in Fig.~\ref{fig_7}).

\subsection{Ablation Study for Sim2Real}

\begin{wraptable}{r}{0.45\textwidth}  
\vspace{-0.4cm}
\centering
\resizebox{\linewidth}{!}{
    \begin{tabular}{cc|c}
        \toprule
        \textbf{DepthAnything} & \textbf{Occupancy} & \textbf{Success Rate (\%) $\uparrow$}  \\ 
        \midrule
        $\checkmark$ & $\checkmark$ & \textbf{86.60} \\
        $\checkmark$ & $\times$    & 73.33  \\
        $\times$   & $\checkmark$ & 60.00  \\
        $\times$   & $\times$ & 33.33  \\
        \bottomrule
    \end{tabular}
}
\caption{Ablation study for sim2real.}
\label{tab:3D-rep}
\vspace{-0.5cm}
\end{wraptable} 
We perform real-world ablation studies to systematically evaluate how individual modules in FetchBot contribute to successful sim-to-real transfer and overall fetching performance. As shown in Table~\ref{tab:3D-rep}, FetchBot w/o DepthAnything significantly reduces the real-world success rate (from 86.60\% to 60.00\%), confirming that predicted depth is crucial for mitigating the visual sim-to-real gap.
Removing the occupancy representation also degrades performance (to 73.33\%), demonstrating its necessity for inferring the occluded regions and alleviating the associated perception limitations.
Using only RGB inputs (removing both) results in the lowest success rate (33.33\%), highlighting the synergistic benefit of the two modules for reliable fetching in challenging real-world conditions.



\section{Conclusion and Limitation}
\textbf{Conclusion.} 
This paper introduces FetchBot, a novel sim-to-real framework tackling the challenge of object fetching in cluttered scenes. By leveraging a large-scale synthetic dataset, depth-conditioned action generation, and techniques to bridge the sim-to-real gap (including RGB-to-depth prediction and occupancy pre-training), FetchBot demonstrates strong zero-shot transfer capabilities. Extensive experiments validate its superior performance in effectively managing clutter and handling challenging objects (e.g., transparent, reflective and irregular), significantly outperforming previous methods and showcasing its practical value for complex, real-world fetching tasks.

\textbf{Limitation.} FetchBot exhibits limitations despite its performance.
Fetchbot exhibits certain limitations when encountering tricky scenarios.
First, in scenarios with significant occlusion, the policy may output complex actions to avoid collisions. However, such maneuvers can cause the robot arm to exceed its joint limits. Second, single-arm grasping of large-volume objects is inherently difficult, suggesting a potential need for dual-arm collaboration in future work. Finally, the challenge with fully occluded targets lies in their initial unreachability. Accessing them requires advanced reasoning to clear obstructions. Ideally, this reasoning would also encompass restoring the scene after the task (e.g., returning moved objects). Developing such comprehensive reasoning capabilities may be a key objective for future work. (Please refer to Appendix~\ref{appendix_f} for more details)

\acknowledgments{
We would like to express our sincere gratitude to Feng Zhu for his invaluable contributions to the real-world experiments, especially for designing the prototype of our suction cup. We are also grateful to Zhennan Jiang, Yupeng Zheng, Jiangran Lv, and Sen Zhang for their insightful discussions and assistance with proofreading. Finally, we would like to thank Fei Wan for his support in rendering the video materials.
}
\clearpage

\end{bibunit}
\clearpage

\appendix
\begin{bibunit}[plain]

\etocdepthtag.toc{supp} 

\begingroup
  \renewcommand{\contentsname}{Appendix} 
  \etocsettagdepth{main}{none}          
  \etocsettagdepth{supp}{subsection}     
  \setcounter{tocdepth}{2}
  \tableofcontents
\endgroup
\clearpage

\appendix
\section{UniVoxGen}
\label{appendix_a}
\subsection{Scene Generation Method}
A diverse and realistic set of scenes is crucial for sim-to-real transfer, requiring an effective scene generation method. Creating a  valid scene involves determining the pose for each object and ensuring there is no penetration among any of them. Therefore, during scene generation, we need to check for collisions between different objects. Previous approaches~\cite{app-wada2022safepicking, app-li2024broadcasting, app-xu2023joint, app-qian2024thinkgrasp,app-yang2024ground4act,app-wang2024learning} for generating cluttered scenes typically depend on computationally intensive collision detection mechanisms. These methods require importing objects into a simulator and executing a simulation step to detect potential collisions, a process known to be computationally expensive. This inefficiency is particularly pronounced in cluttered scenes, where the dense packing of objects inherently leads to frequent collisions requiring verification. Additionally, some other methods create cluttered layouts by simulating objects dropped from the air, which may result in unstable and unrealistic scene configurations.
Our method, UniVoxGen, is specifically designed for fast and realistic scene generation in voxel space. It accelerates the generation process by performing efficient collision checks in voxel space and produces realistic scene layouts using a set of carefully crafted hand-designed rules.

We begin by providing formal definitions for key elements in the voxel space. Let $V^o = \{V^o_{1}, V^o_{2}, \dots, V^o_{N}\}$ represent the voxel representation of a set of objects, and $V^s = \{V^s_{1}, V^s_{2}, \dots, V^s_{N}\}$ represent the voxel representation of the scene. We define a set of operational primitives in voxel space for manipulating voxels. Specifically, $V_i \cup V_j$ denotes the union operation, which combines two voxel sets and is commonly used to add an object’s voxels into the scene. $V_i \cap V_j$ denotes the intersection operation, which retrieves the intersection of two voxel sets and is used to detect potential collisions when adding a new object. $V_i - V_j$ denotes the difference operation, which removes the overlapping portion of $V_i$ with $V_j$, typically used to remove an object from the scene. 
Finally, $T(V_i, P), P \in {SE}(3)$ represents a transformation of a voxel $V_i$ in ${SE}(3)$ space, commonly used to change the pose of the object. Here, $P$ is a transformation matrix in ${SE(3)}$ that combines rotation and translation.

\begin{algorithm}
\caption{Scene Generation Algorithm}
\begin{algorithmic}[1]
\REQUIRE Number of scenes $N$
\REQUIRE Max objects per scene $K$
\REQUIRE A set of objects $\mathcal{O}$
\FOR{scene $1:N$}
    \STATE Initialize scene voxel $V^s = \{\}$
    \STATE Sample a target object $O^{tar}$
    \STATE Sample a pose $P$ in SE(3) for $O^{tar}$
    \STATE Apply $T(V_i, P)$ to transform the target object
    \STATE Apply $V_{O^{tar}} \cup V^s$ to add the target object to the scene
    \STATE Sample number of obstacle objects $k \sim [1, \ldots, K]$
    \FOR{obstacle $O^{obs}_{i}$ $1:k$}
        \STATE Sample a pose $P$ in SE(3) for $O^{obs}_{i}$
        \STATE Apply $T(V_i, P)$ to transform the obstacle object
        \WHILE{$V_{O^{obs}_{i}} \cap V^s$}
            \STATE Sample a new pose $P$ in SE(3) for $O^{obs}_{i}$
            \STATE Apply $T(V_i, P)$ to transform the obstacle object
        \ENDWHILE
        \STATE Apply $V_{O^{obs}_{i}} \cup V^s$ to add the obstacle object to the scene
    \ENDFOR
    \STATE Save the pose $P$ of each object in the scene
\ENDFOR
\end{algorithmic}
\label{alg:scene_gen}
\end{algorithm}

Based on the previously defined key elements and operation primitives, we further design a set of generation rules $R = \{R_1, R_2, \dots, R_N\}$. UniVoxGen uses these rules to generate diverse cluttered scenes. These scenes may include unsolvable cases, where it is impossible to retrieve the target object without colliding with any obstacles. It is worth noting that the inclusion of unsolvable cases is intended to better simulate real-world scenarios, as such situations can occur in practice. The procedure for generating these cluttered scenes is outlined in Algorithm~\ref{alg:scene_gen}. It should be noted that, given the complexity of the various scene generation rules, the steps presented here represent a simplified version of our scene generation rules. The detailed generation rules will be made available in the released source code. Finally, we use UniVoxGen to generate \textbf{\textit{1 million}} cluttered scenes, which are then utilized as training scene data for a oracle policy. It takes 12 hours on the workstation equipped with 8 RTX 4090s to generate these 1 million scenes.

\subsection{Scene Generation Result}
\begin{figure*}[tb]
    \centering
    \includegraphics[trim=3.2cm 6cm 6cm 0.0cm, clip,  width=1.0\linewidth]{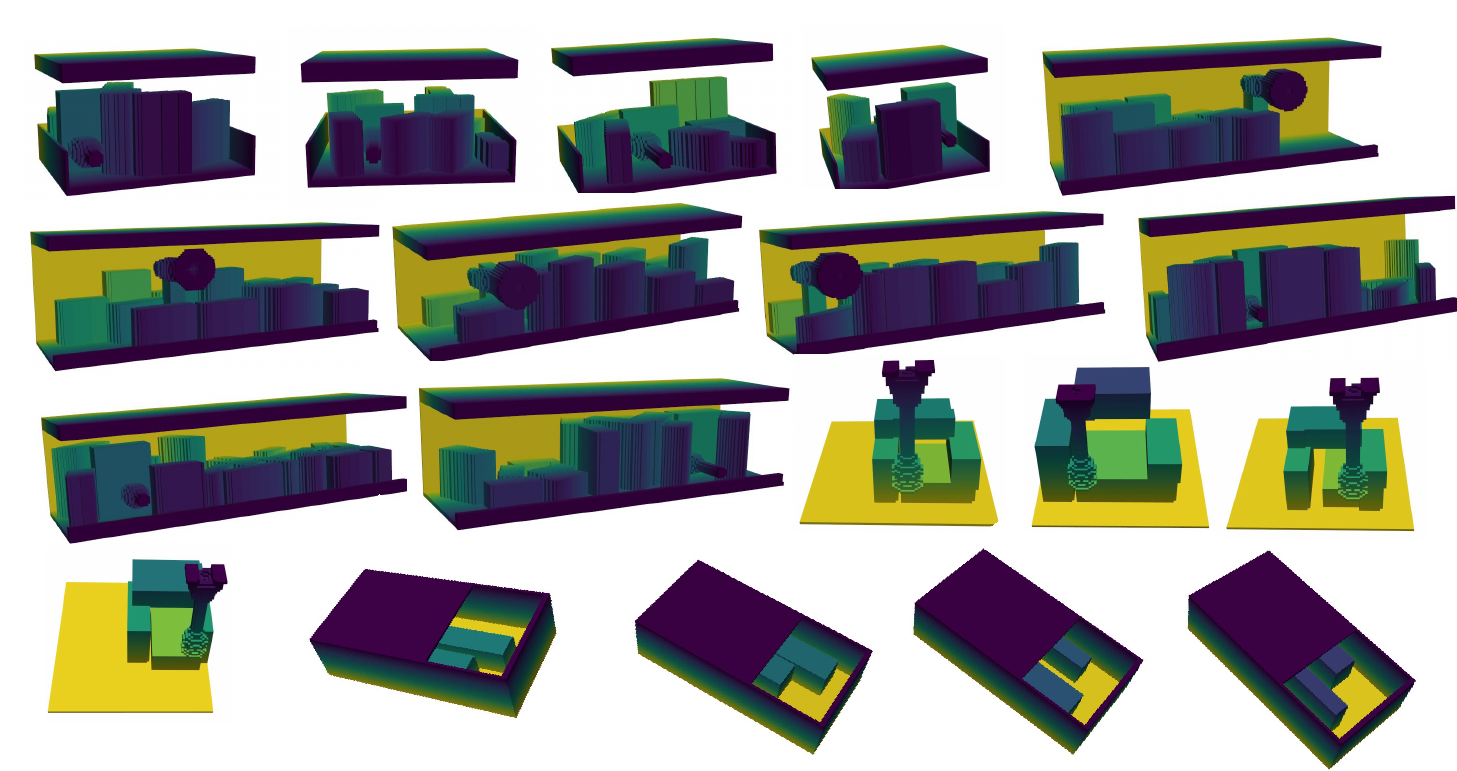}
    \caption{Generated cluttered scenes by UniVoxGen, including shelf, tabletop, drawer, and storage rack environments.}
    \vspace{-1em}
    \label{fig:univoxgen}
\end{figure*}

We utilize UniVoxGen to generate a large number of cluttered environments. Our initial focus is on cluttered shelf environments. Such environments require not only clutter but also a certain degree of structured arrangement to be realistic. On a shelf, objects cannot be arbitrarily piled together, but instead need to exhibit some regularity in their placement to simulate realistic shelf settings. Additionally, we generate cluttered environments for tabletop, drawer, and rack settings to support the needs of extension experiments, as shown in Fig.~\ref{fig:univoxgen}.

\section{Oracle Policy Training}
\label{appendix_b}
\subsection{Reward Function}
Our reward function combines task completion $r_{task}$, behavioral constraints $r_{cons}$, and an environment change penalty $r_{env}$.

\textbf{Task Reward.} 
The task reward formulation imposes a dual requirement: the target object must be successfully fetched, and the resulting environmental impact, measured over the entire episode, must be limited. This impact is quantified by the total translation ($m_{trans}$) and total rotation ($m_{rot}$) accumulated across all obstacles during the episode. Both metrics must remain below their respective predefined thresholds: $m_{trans}$ must be less than the translation threshold $\sigma_{trans}$, and $m_{rot}$ must be less than the rotation threshold $\sigma_{rot}$. We use a curriculum learning method for the thresholds ($\sigma_{trans}$ and $\sigma_{rot}$), gradually decreasing their values during training. Specifically, $\sigma_{trans}$ is set according to the sequence [0.03, 0.015, 0.01, 0.005, 0.0] m, and $\sigma_{rot}$ follows the sequence [0.4, 0.2, 0.16, 0.1, 0.0] radians. The transition to the next value in each sequence occurs once the policy's performance saturate at the current stage.

\textbf{Behavioral Constraints Reward.} 
We apply certain behavioral constraints to the flying end-effector to make its behavior more naturalistic.

We use $r_{action\_range}$ to constrain the policy's output from becoming too large. That is: $$
r_{action\_range} =
\begin{cases}
    -\lambda_{action\_range} \exp\left( (|a| - 3)^2 \right) & \text{if } |a| > 3 \\
    0 & \text{if } |a| \le 3
\end{cases}
$$

To ensure smoother and more continuous actions, we introduce a term, $r_{action\_rate}$, that penalizes large differences between consecutive pose outputs of the policy ($a_{last}, a_{current}$):$$
r_{action\_rate} = -\lambda_{action\_rate} \exp\left( (a_{last} - a_{current})^2 \right)
$$

To prevent fetching the object in singular poses, which could violate robot arm joint limits, we incentivize maintaining the target object close to its initial pose ($p_{init}$). This objective is achieved using the $r_{pose}$ penalty term, calculated as:
$$
r_{pose} = -\lambda_{pose} \exp\left( p_{curr} - p_{init} \right)
$$
  
To prevent collisions with shelf barriers, we introduce a penalty term, $r_{penetration}$. If a collision, i.e., penetration, occurs, the target object experiences a significant acceleration, denoted as $a_{ccel}$. This acceleration is then used to compute the penalty as follows:
$$
r_{penetration} = -\lambda_{pen} \exp\left( a_{ccel}^2 \right)
$$

To encourage moving the target towards its designated goal position, we employ a reward component, $r_{target\_move}$, to guide the policy's behavior. This reward incentivizes movement towards the goal and penalizes movement away from it. Let $d_{last}$ denote the distance from the target to the goal position at the previous timestep, and $d_{curr}$  be the current distance to the goal. The term $d_{last} - d_{curr}$ thus represents the progress made towards the goal in the current step, a positive value indicates movement closer to the goal. We use $v$ represents target's velocity and $dt$ is the duration of the timestep, the reward is defined as:
$$
r_{target\_move} = \lambda_{target\_move} \cdot \frac{d_{last} - d_{curr}}{v \cdot dt}
$$

\textbf{Environment Change Reward.}
To incentivize the policy to minimize its impact on the environment, we introduce penalties based on the total displacement of all obstacles within each timestep. Specifically, we consider the total translational displacement ($m_{trans\_step}$) and the total rotational displacement ($m_{rot\_step}$) of all obstacles. These penalties are computed as follows:
$$
r_{trans\_step} = -\lambda_{trans\_step} \exp\left( m_{trans\_step}^2 \right)
$$
$$
r_{rot\_step} = -\lambda_{rot\_step} \exp\left( m_{rot\_step}^2 \right)
$$
All weight coefficients are listed in Table~\ref{tab:rewards}. We train our oracle policy with PPO~\cite{app-schulman2017proximal}, and the training hyper-parameters are shown in Table~\ref{tab:rl}.


    


    

\begin{table}[htbp]
\centering
\begin{minipage}[t]{0.48\linewidth}
    \centering
    \resizebox{0.9\linewidth}{!}{
        \begin{tabular}{l|c} 
        \toprule
        Hyper-parameters         & Values    \\ 
        \midrule
        \textit{Num. envs}     &      \textit{1024}        \\
        \textit{Num. steps for per update}     &      \textit{24}        \\
        \textit{Num. minibatches}     &      4        \\
        \textit{Num. learning epochs}     &      \textit{1500}        \\
        \textit{learning rate}     &      \textit{0.0003}        \\
        \textit{clip range}     &     \textit{ 0.2}        \\
        \textit{entropy coefficient}     &      \textit{0.0}        \\
        \textit{kl threshold}     &      \textit{0.02}        \\
        \textit{max gradient norm}     &     \textit{1.0}        \\
        $\lambda$   & \textit{0.95}\\
        $\gamma$ & \textit{0.99}\\
        \bottomrule
        \end{tabular}
    }
    \vspace{0.5em}
    \caption{Hyper-parameters for the oracle policy learning.}
    \label{tab:rl}
\end{minipage}
\hfill
\begin{minipage}[t]{0.48\linewidth}
    \centering
    \resizebox{0.75\linewidth}{!}{
        \begin{tabular}{l|c} 
        \toprule
        Hyper-parameters         & Values    \\ 
        \midrule
        $\lambda_{task}$     &      \textit{5.0}        \\
        $\lambda_{action\_range}$     &     \textit{ 1.5}       \\
        $\lambda_{action\_rate}$     &      \textit{0.0001}        \\
        $\lambda_{pose}$     &      \textit{0.6 }      \\
        $\lambda_{penetration}$     &     \textit{ 9.0 }       \\
        $\lambda_{target\_move}$     &     \textit{ 0.06 }      \\
        $\lambda_{trans\_step}$     &     \textit{ 10.0 }       \\
        $\lambda_{rot\_step}$     &     \textit{ 10.0 }     \\
        \bottomrule
        \end{tabular}
    }
    \vspace{0.5em}
    \caption{Hyper-parameters for the reward function.}
    \label{tab:rewards}
\end{minipage}
\end{table}

\subsection{Training Randomization}
To improve the robustness of the oracle policy and diversify the demonstration set, we apply extensive domain randomization during training. We randomize the objects’ mass and center of mass, friction coefficients, the shelf’s upper-barrier height, camera pose, and the end-effector’s initial pose; in addition, we inject noise into observations to emulate real-world imperfections such as calibration bias and readout noise.

\subsection{Curriculum Learning Methods}
We explore various curriculum learning strategies. The first step is to assign difficulty levels to the task scenarios. Specifically, we generate a diverse set of scenarios based on predefined rules, and then categorize them into five difficulty levels according to the occlusion rate—defined as the percentage of the target object’s surface area occluded by obstacles when viewed from the front.

\textbf{\textcolor{myred}{Approach 1. \textit{Per-Environment Difficulty Curriculum.} \ding{55}}} We adopt a locomotion-style curriculum~\cite{app-zhuang2023robot} as our first attempt: in Isaac Gym, each environment is assigned a difficulty level; upon task success its difficulty increases, otherwise it decreases. However, this scheme does not yield any improvement over direct training on all scenarios.

\textbf{\textcolor{myred}{Approach 2. \textit{Unified Difficulty Curriculum.} \ding{55}}} Our second attempt departs from the locomotion-style curriculum. In Isaac Gym, all parallel environments are kept at the same difficulty level, rather than assigning each environment its own difficulty. Once the policy converges at the current difficulty, we advance all environments to the next level. However, this method also fails to produce satisfactory results.

\textbf{\textcolor{myred}{Approach 3. \textit{Policy-Driven Difficulty Curriculum.} \ding{55}}} Our third attempt abandons the use of manually defined metrics (i.e., occlusion rate) for defining difficulty levels, rather than relying on the policy’s own performance. Specifically, we first train the policy on all scenarios until convergence, then evaluate it and label the failed scenarios as higher-difficulty cases. Training then continues exclusively on these failed scenarios, and this process is repeated for five iterations. However, this method also fails to deliver satisfactory results.

\textbf{\textcolor{mygreen}{Approach 4. \textit{Task-Conditioned Curriculum.} \ding{51}}} Our final method is a $\sigma$-based curriculum (\S~\ref{oracle_policy}): $\sigma$ is large early for exploration and is gradually reduced to enhance precision and stability.

Our key insight is that, whether difficulty is defined manually or in a policy-driven way, the chosen scene difficulty axis is highly homogeneous and lacks qualitative transitions across levels; knowledge learned in easy scenarios transfers poorly to hard ones, failing to provide effective learning gradients or a skill ladder. Unlike scene-based curricula, the $\sigma$-based curriculum decouples training from scene parameters and schedules solely based on task difficulty via $\sigma$, thereby avoiding scene-difficulty specification and directly improving task performance and training stability.

\subsection{The Architecture of Oracle Policy}
\textbf{Encoder.} The oracle policy receives an observation composed of three parts:
the scene representation \( \boldsymbol{z}_t \), proprioception \( \boldsymbol{p}_t \),
and the previous action \( \boldsymbol{a}_{t-1} \).
The term \( \boldsymbol{p}_t \in \mathbb{R}^{12} \) encodes the end-effector pose
relative to the target placement and decomposes as
\( \boldsymbol{p}_t = [\mathrm{rot}_t,\, \mathrm{trans}_t] \),
where \( \mathrm{rot}_t \in \mathbb{R}^{9} \) denotes the rotation matrix
(flattened to 9 entries) and \( \mathrm{trans}_t \in \mathbb{R}^{3} \) is the
translation vector. The previous action satisfies
\( \boldsymbol{a}_{t-1} \in \mathbb{R}^{6} \).
The scene encoder network (\S~\ref{oracle_policy}) maps the raw scene input to
\( \boldsymbol{z}_t \in \mathbb{R}^{64} \).
The final observation is the concatenation \( \boldsymbol{o}_t = [\,\boldsymbol{z}_t,\,\boldsymbol{p}_t,\,\boldsymbol{a}_{t-1}\,] \in \mathbb{R}^{82} \).

\textbf{Decision.} The decision module is a three-layer MLP that maps the
observation to an action. Formally, the policy \( \boldsymbol{\pi} \) implements
\( \boldsymbol{\pi} : \mathbb{R}^{82} \!\to\! \mathbb{R}^6 \) and
\( \boldsymbol{a}_t = \boldsymbol{\pi}(\boldsymbol{o}_t) \), where \( \boldsymbol{a}_t \in \mathbb{R}^6 \).

\section{Vision Policy Training}
\label{appendix_c}
\subsection{Ablation Study on Scaling the Training Data}
\begin{figure}[htbp]
    \centering
   
    \begin{minipage}[t]{0.48\textwidth}
        \centering
        \includegraphics[trim=0cm 0cm 0cm 0.0cm, clip, width=\linewidth]{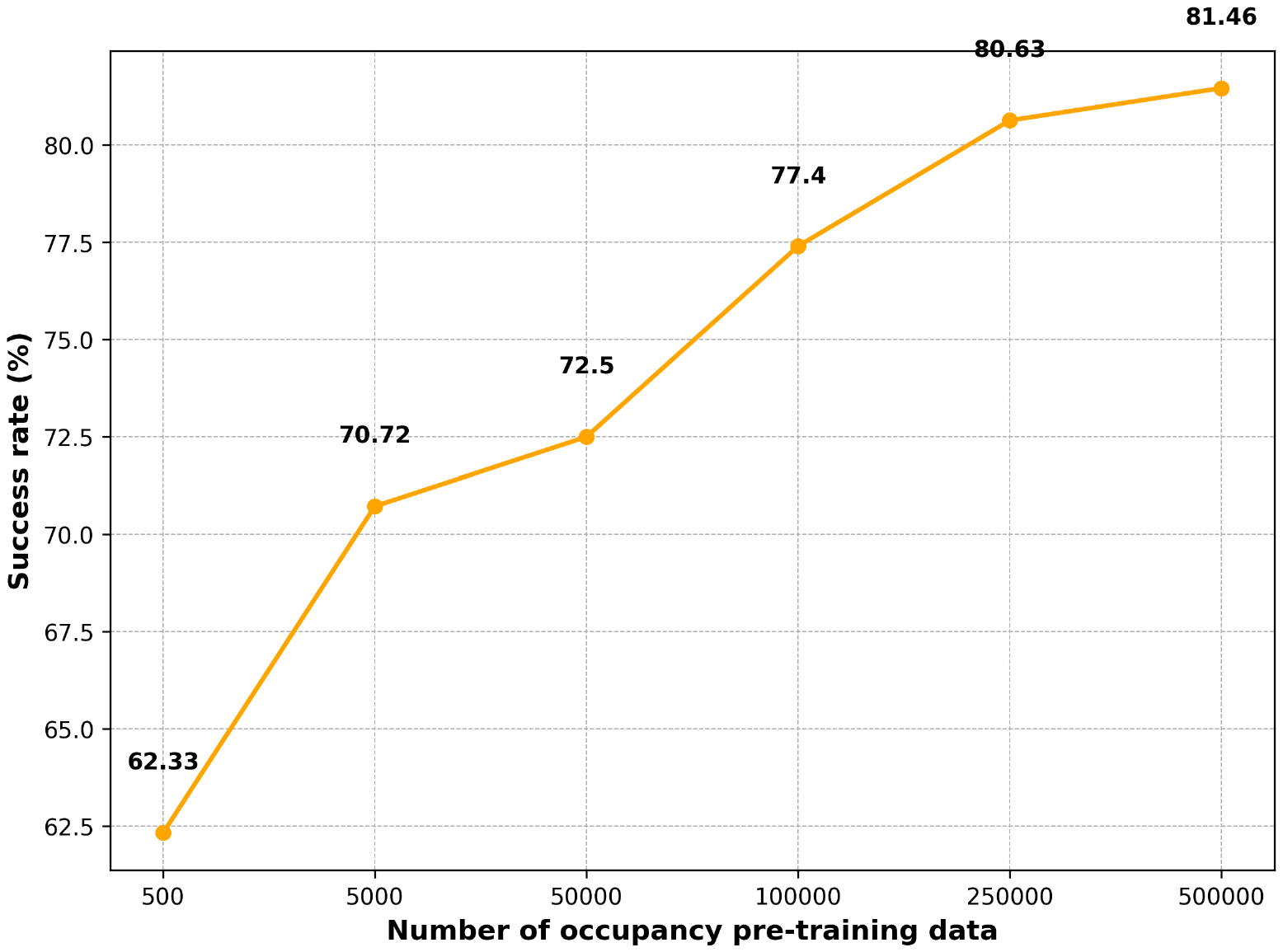}
        \caption{Occ Pre-Training Scaling Law}
        \label{fig:occ_scaling_law}
    \end{minipage}
    \hfill
     \begin{minipage}[t]{0.48\textwidth}
        \centering
        \includegraphics[trim=0cm 0cm 0cm 0.0cm, clip, width=\linewidth]{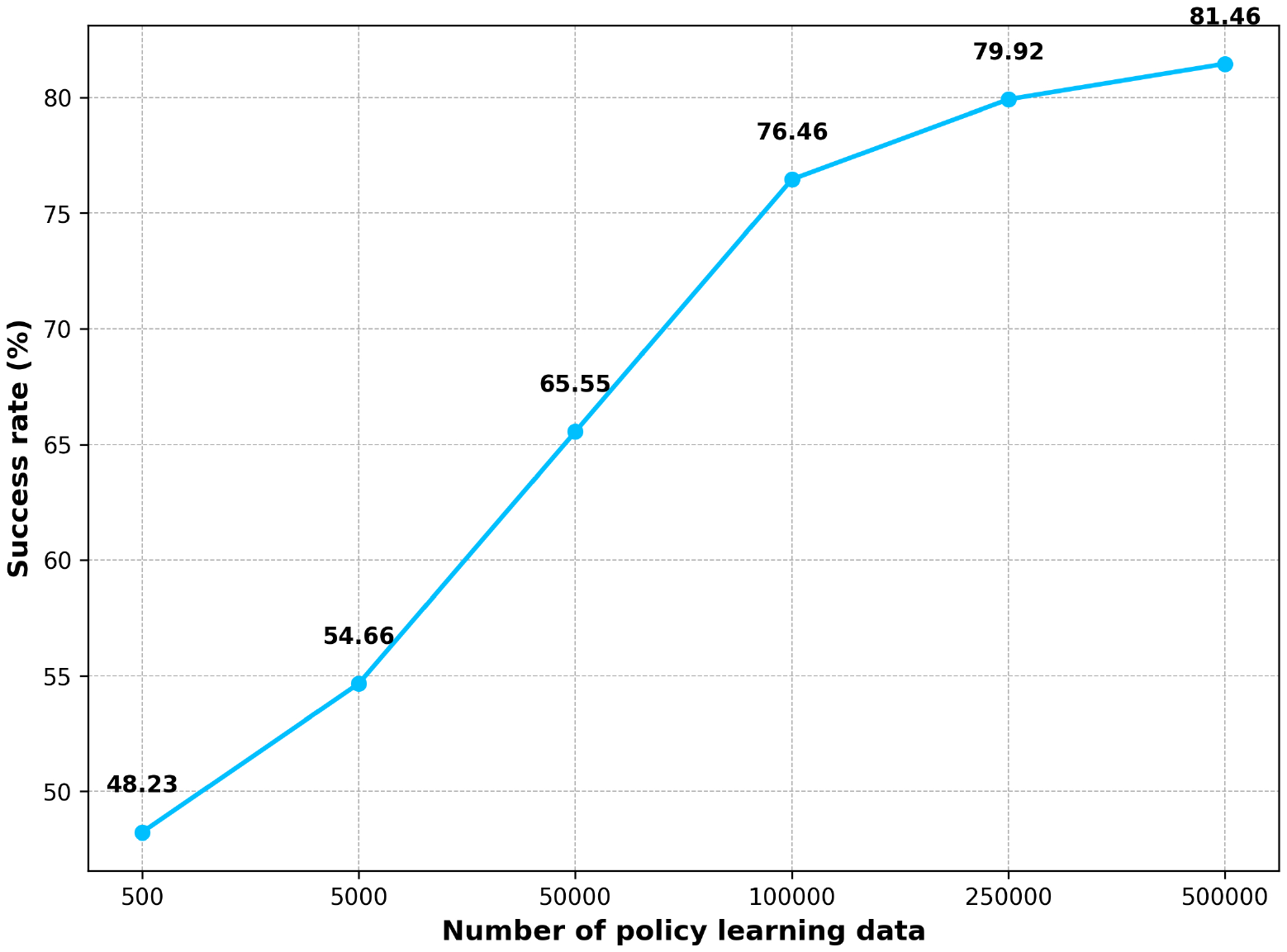}
        \caption{Policy Learning Scaling Law}
        \label{fig:policy_scaling_law}
    \end{minipage}
\end{figure}

In our ablation study on scaling the training data, we examine its impact on both the occupancy pre-training (Fig.~\ref{fig:occ_scaling_law}) and the policy learning (Fig.~\ref{fig:policy_scaling_law}). We use suction cup as the end-effector tool to complete this experiment.

\textbf{Scaling the Training Data for Occupancy Pre-Training.} 
This study examines how the amount of data utilized for pre-training the 3D vision encoder via an occupancy prediction task affects the ultimate performance of the model. All experiments employ varying dataset sizes for pre-training the 3D vision encoder but maintain a fixed dataset size of 500K for policy training.
We find a clear correlation between larger datasets and improved policy performance. Starting with 500 scenes, the success rate is 62.33\%. As the dataset size increases to 5k and 50k, the success rate improves to 70.72\% and 72.50\%, respectively. 
As the dataset size increases to 100k and 250k, the performance continues to improve.
The largest dataset, with 500k scenes, achieves the best performance, reaching a success rate of 81.46\%. 
This demonstrates that pre-training on larger datasets significantly enhances the policy performance, providing a more comprehensive understanding of the 3D scene.

\textbf{Scaling the Training Data for Policy Learning.}
We seek to determine how the quantity of data available for policy learning influences the model's ultimate performance. For this investigation, all experiments maintain a fixed data volume for 3D vision encoder pre-training but use different amounts of data for training the policy.
With a dataset of 500 state-action pairs, the success rate is 48.23\%. As the dataset increases to 5k and 50k, the success rate rises to 54.66\% and 65.55\%, respectively. 
As the dataset size increases to 100k and 250k, performance continues to improve.
The largest dataset of 500k state-action pairs yields the best performance, achieving a success rate of 81.46\%. These results highlight that increasing the training data size for the decision module significantly improves task success, emphasizing the importance of a sufficiently large dataset for accurate and reliable policy performance.

Overall, our findings highlight the critical role of large-scale data in both occupancy pre-training and policy learning, underscoring the importance of scaling the dataset for improving the overall performance and accuracy of the 3D vision policy.

\subsection{Ablation Study on Region of Interest (ROI) Size}

Focusing solely on the region of interest is an effective approach that enhances the policy’s generalization ability and aids policy learning. We define our Region of Interest (ROI) by cropping an $H \times W \times Z$ cm cubic space around the end-effector.

In our ablation study on the ROI size (Table.~\ref{tab:roi}), we investigated the impact of varying sizes on policy performance. The results show that a compact ROI, specifically $20 \times 20 \times 30$ cm, leads to the best performance, achieving the highest success rate (81.46\%) and the lowest translation (2.78 cm) and rotation (0.36 rad) errors. It is crucial to note that \textbf{\textit{the optimal ROI size is not arbitrarily small but is closely tied to the scale of the target objects.}} In our experiments, the target objects average approximately $5\times 10 \times 12$ cm. Therefore, an ROI of $20 \times 20 \times 30$ cm is large enough to fully encompass the target while providing essential local context, yet small enough to filter out most distracting information. As the ROI size increases beyond this optimal range, both the success rate and accuracy metrics decrease. This is because larger ROIs introduce more irrelevant information from distant areas, which reduces the model's ability to focus on the target region and leads to less precise predictions.

These findings highlight the advantage of using a \textbf{\textit{task-appropriate, localized ROI.}} This approach not only improves policy performance but also enhances generalization, making the network more robust to scene variations. It also accelerates policy inference while maintaining high performance, especially in cluttered environments.

\begin{table}[htbp]
\centering
\resizebox{0.7\linewidth}{!}{
    \begin{tabular}{l|ccc}
    \toprule
    \textbf{ROI Size (cm)} & \textbf{Success Rate (\%) $\uparrow$} & \textbf{Translation (cm)$\downarrow$} & \textbf{Rotation (rad)$\downarrow$} \\
    \midrule
     $20\times 20\times 30$                &         \textbf{81.46\%}        &  \textbf{2.78}         &   \textbf{0.36}                        \\
    $40\times 80\times 30$                &       76.93\%           &       3.24                   &        0.43                 \\
    $60\times 150\times 30$               &      74.63\%            &       3.67                     &      0.47                   \\
    \bottomrule

    \end{tabular}
}
\vspace{0.7em}
\caption{ 
\textbf{Ablation study on Region of Interest (ROI) size.} A compact ROI of $20 \times 20 \times 30$ cm achieves the best results, with the highest success rate and lowest errors. This highlights the benefit of a localized ROI that is appropriately scaled to the task, as larger ROIs degrade performance by including irrelevant information from distant areas.
}
\vspace{-2em}
\label{tab:roi}
\end{table}

\subsection{Training Details}
Pre-training for occupancy prediction used 40 GeForce RTX 4090 GPUs on 500K scenes and finished in 16 hours. Policy learning then trained on 500K demonstration frames for 8 hours. In real-world tests, the system runs at over 10 fps on a single NVIDIA GeForce RTX 4090.

\subsection{The Details of 2D-3D Spatial Attention}
Rather than averaging multi-view features (which assumes equal contribution), we fuse them using Deformable Cross-Attention (DCA)~\cite{app-zhu2020deformable}, guided by a grid of 3D queries $Q \in \mathbb{R}^{C \times H \times W \times Z}$.

As shown in Fig.~\ref{fig_real_compare_voxel_occ}, for each 3D query, defined by its position $p$ and feature $q_p$, we project it onto each camera view $t$ via $\mathcal{P}(p, t): \mathbb{R}^3 \mapsto \mathbb{R}^2$. We retain only the set of views where the projection is valid, $\mathcal{V}_{\text{hit}}$. The aggregated 3D feature $\mathcal{F}^p$ at voxel $p$ is then computed by averaging the attention outputs from all visible views:
\[
\mathcal{F}^p = \mathrm{DCA}(q_p, \{\mathcal{F}^{2D}_t\}) = \frac{1}{|\mathcal{V}_{\text{hit}}|} \sum_{t \in \mathcal{V}_{\text{hit}}} \mathrm{DA}\big(q_p, \mathcal{P}(p, t), \mathcal{F}^{2D}_t\big),
\]
where $\mathcal{F}^{2D}_t$ is the 2D feature map of view $t$. The Deformable Attention (DA) module is defined as:
\[
\mathrm{DA}(q, p_{\text{ref}}, X) = \sum_{m=1}^{N_{\text{head}}} W_m \left( \sum_{k=1}^{N_{\text{key}}} A_{mk} \cdot W'_m X(p_{\text{ref}} + \Delta p_{mk}) \right).
\]
Here, for each attention head $m$ and key $k$, the module uses the query $q$ to predict a 2D sampling offset $\Delta p_{mk}$ relative to the reference point $p_{\text{ref}} = \mathcal{P}(p,t)$ and an attention weight $A_{mk}$. It then samples the 2D feature map $X$ at these sparse, dynamically determined locations ($p_{\text{ref}} + \Delta p_{mk}$) and computes a weighted sum. $W_m$ and $W'_m$ are standard learnable projection matrices.

This approach allows the final 3D feature volume, $\mathcal{F}=\{\mathcal{F}^p\}$, to unequally weight views based on visibility and content, enhancing robustness to occlusion and blur.

\subsection{Processing of Relative Depth}
Depth estimates from DepthAnything~\cite{app-yang2024depth} are inherently scale-ambiguous. To create a consistent input for our feature extractor, we first normalize each predicted depth map using a per-image \textbf{min-max scaling} to bring its values into the range $[0, 1]$.

Concretely, given an RGB image $I \in \mathbb{R}^{H \times W \times 3}$, DepthAnything produces a single-channel depth map $D \in \mathbb{R}^{H \times W}$. This map is normalized as:
\[
\hat{D} = \frac{D - \min(D)}{\max(D) - \min(D) + \varepsilon}
\]
where $\min(D)$ and $\max(D)$ are the minimum and maximum values of the specific depth map $D$, and $\varepsilon$ is a small constant for numerical stability.

To prepare this normalized map for a standard ResNet backbone, we treat $\hat{D}$ as a grayscale image. It is first replicated across three channels to match the expected input format ($H \times W \times 3$). Then, this 3-channel tensor undergoes a second normalization using the standard ImageNet mean and standard deviation values. The resulting tensor is finally fed into the ResNet backbone (with the final max-pooling layer removed) to extract the feature map $F$.

\subsubsection{Mapping Relative Depth to Metric Voxels}
Our method does not perform an explicit conversion from relative depth to metric depth, nor does it use depth values to unproject 2D features into a 3D point cloud. Instead, the core of our approach is to project the centers of a predefined 3D metric voxel grid (with a cell size of 5 mm) onto the input camera views. This forward projection is based on the standard pinhole camera model and utilizes the provided camera matrices.

This multi-view projection strategy is key to resolving spatial ambiguity. By using multiple views, points that might lie along a single viewing ray from one camera are disambiguated through parallax, as they project to distinct 2D locations in the second view.

Consequently, the network is tasked with implicitly learning the transformation from relative depth features to a metrically-scaled occupancy grid, guided by these strong multi-view geometric constraints. The entire process is optimized end-to-end with direct supervision from the ground-truth metric voxel data. In summary, our methodology leverages the geometric accuracy of the pinhole camera model for projection and the power of end-to-end learning to fuse features from relative depth into a usable metric representation.

\subsection{Real-World Reconstruction Results}
\begin{figure*}[tb]
    \centering
    \includegraphics[trim=0.0cm 0.1cm 2.0cm 0.0cm, clip,  width=1.0\linewidth]{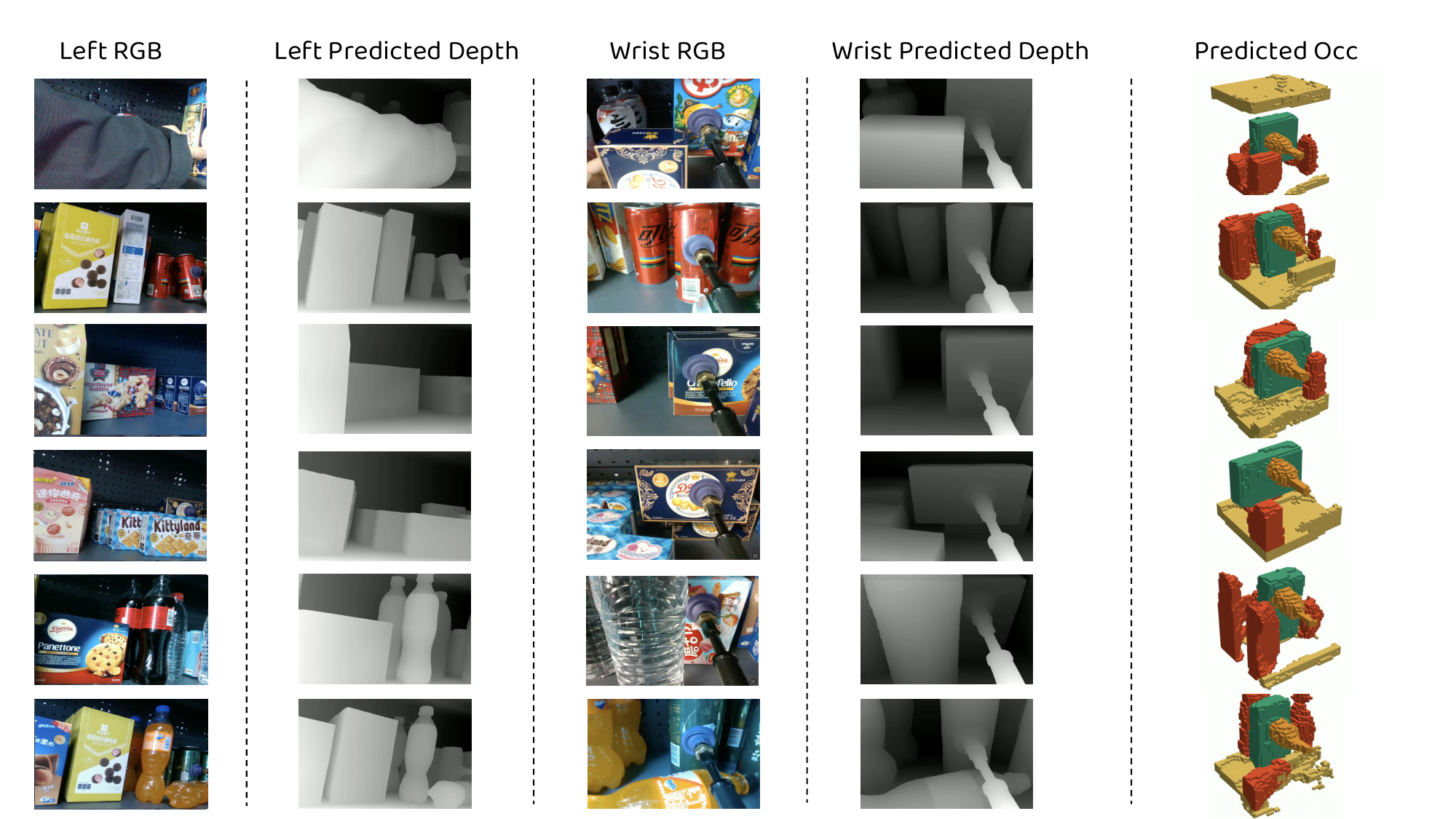}
    \caption{Real-world occupancy reconstruction results, capable of handling diverse scenarios: varying object shapes, different materials, and complex layouts.}
    \label{fig:real_world_occ}
    \vspace{-1em}
\end{figure*}

We first pre-train the encoder on an occupancy-prediction auxiliary task in simulation, which yields high-quality reconstruction not only in simulation but also on real hardware (strong sim-to-real, as shown in Fig.~\ref{fig:real_world_occ}). Under heavy occlusions, the model infers the occluded volume to recover a complete scene representation, and it remains robust to transparent, reflective, and irregular objects. Importantly, we do not feed the reconstructed voxels (final occupancy map) to the downstream policy. Instead, the policy receives intermediate latent features from the pre-trained encoder; although it never observes the final occupancy prediction, these auxiliary-task–enriched features allow it to implicitly capture the scene’s complete 3D geometry.
\section{Baseline Implementation Details}
\label{appendix_d}
Baselines utilizing RGB (Diffusion Policy), point cloud (3D Diffusion Policy), RGB-D voxels, raw depth, and predicted depth all share the same action generation method, differing only in their input representations. Actions are generated via a Denoising Diffusion Probabilistic Model (DDPM)~\cite{app-ho2020denoising}, which utilizes 1000 denoising steps during training and 100 steps for inference, its network architecture is a three layer MLP.

\textbf{RGB (Diffusion Policy)}~\cite{app-chi2023diffusion}.
We use dual-view RGB images as input and extract features following the same processing pipeline as Diffusion Policy (i.e., through ResNet and Spatial Softmax). Subsequently, actions are generated using DDPM.
\begin{figure*}[tb]
\centering
\begin{minipage}[t]{0.46\linewidth}
    \centering
    \includegraphics[trim=0cm 6.5cm 13cm 0.0cm, clip, width=1.1\linewidth]{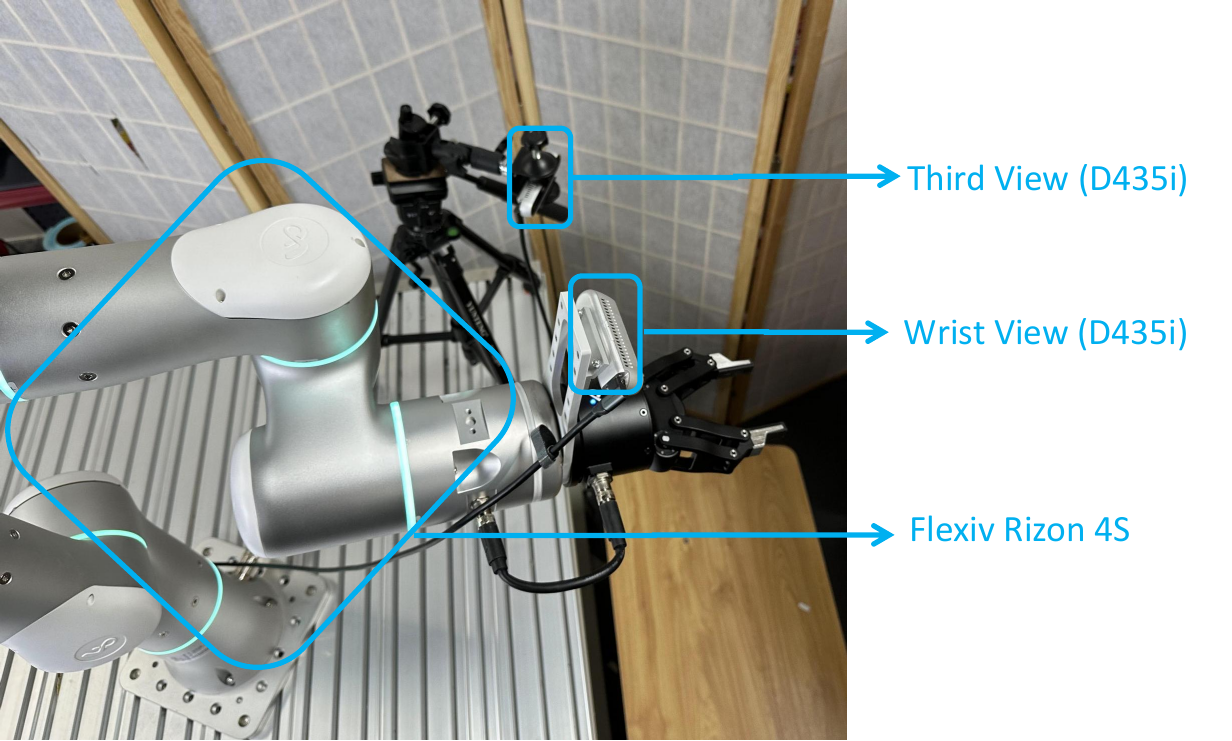}
    \caption*{(a) Hardware Setup}
    \label{fig:hardware}
\end{minipage}
\hspace{0.5cm}
\begin{minipage}[t]{0.48\linewidth}
    \centering
    \includegraphics[trim=0.3cm 8cm 20cm 0.0cm, clip, width=0.9\linewidth]{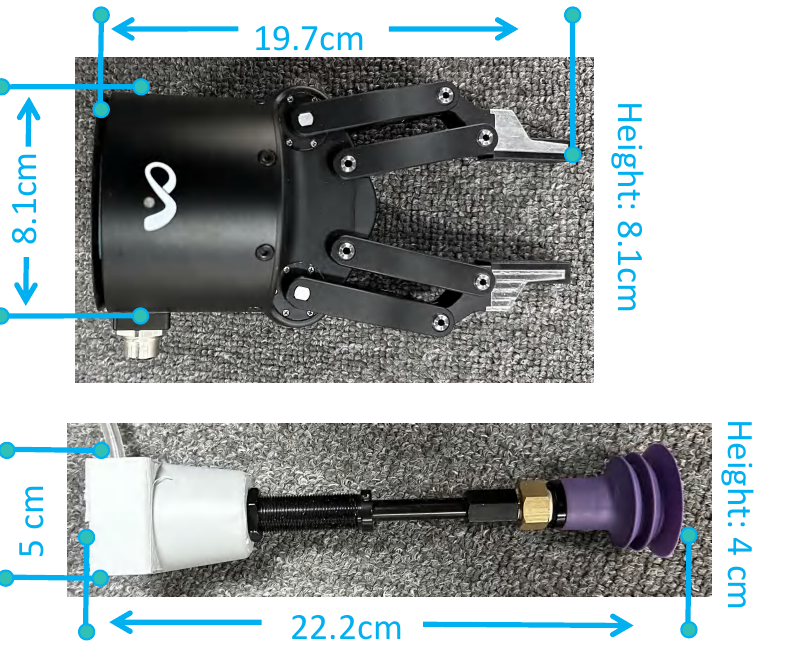}
    \caption*{(b) End-Effector Tools}
    \label{fig:tools}
\end{minipage}
\vspace{0.5em}
\caption{We illustrate our real-robot experimental hardware setup and the two types of end-effector tools employed: one suction cup and one parallel gripper.}
\label{fig:hardware_tools}
\end{figure*}

\textbf{Point Cloud (3D Diffusion Policy)}~\cite{app-ze20243d}.
Our input is a point cloud generated by fusing two camera depth maps. Following the approach in DP3, we employ a Simple PointNet to extract features from this point cloud.

\textbf{RGB-D Voxel.}
We project RGB images from two viewpoints into 3D space using camera extrinsics and depth maps, thereby forming a voxel representation. A 3D convolutional network (3D ConvNet) is then employed to extract features from these voxels. Then use DDPM to generate actions.

\textbf{Raw Depth.}
We use ResNet to extract features from two viewpoint depth maps, after which actions are generated via DDPM.

\textbf{Predicted Depth.}
We utilize Depth Anything to obtain predicted depth maps. Features are then extracted from these maps using ResNet. Subsequently, actions are generated via DDPM.
\begin{figure*}[tb]
    \centering
    \includegraphics[trim=0.0cm 9cm 7cm 0.0cm, clip,  width=1.0\linewidth]{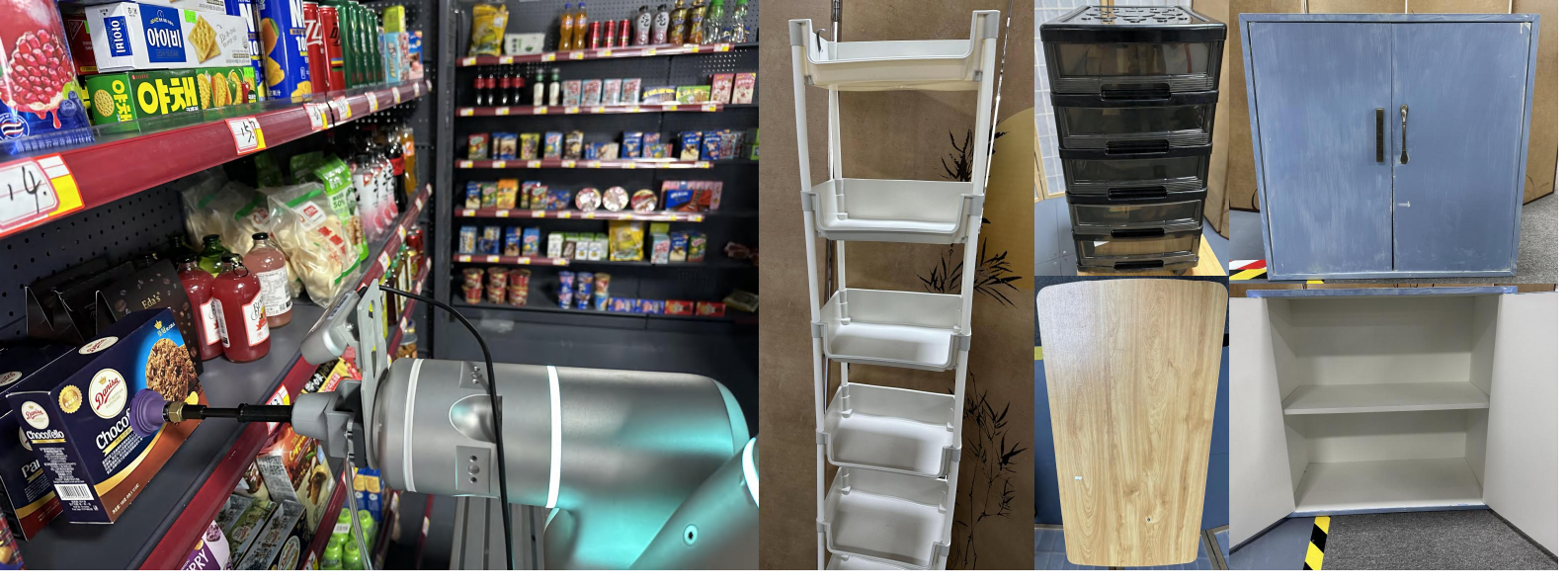}
    \caption{Illustration of experimental scenarios. The main experiments are conducted in a replicated retail store, supplemented by evaluations on a storage rack, in a cabinet, in a drawer, and on tabletops.}
    \vspace{-1em}
    \label{fig:scenes}
\end{figure*}

\begin{figure*}[tb]
    \centering
    \includegraphics[trim=0.0cm 5.5cm 12cm 0.0cm, clip,  width=1.0\linewidth]{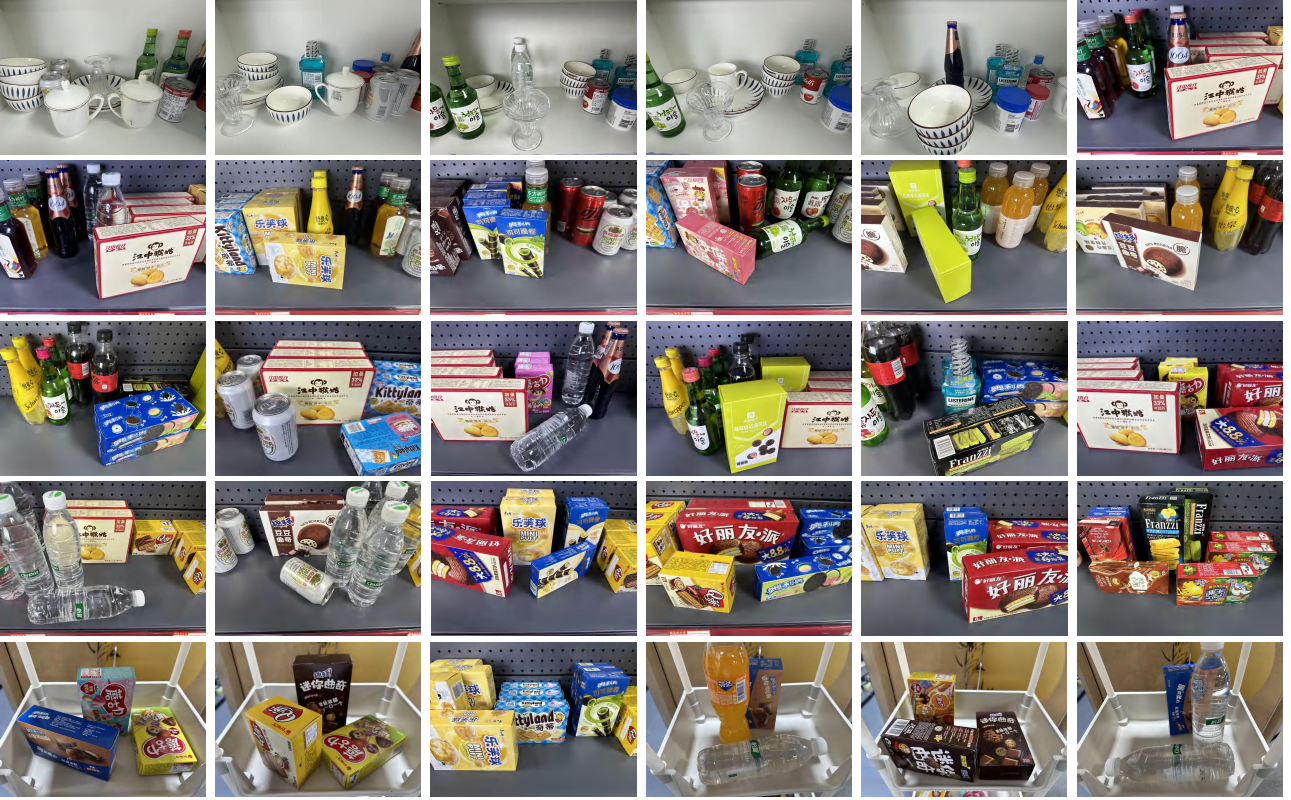}
    \caption{Layouts for real-robot experiments.}
    \label{fig:layouts}
    \vspace{-1em}
\end{figure*}
\section{Real-Wold Experiment Details}
\label{appendix_e}
\subsection{Setup}
\textbf{Hardware.}
In our real-world experiments, we use the Flexiv Rizon 4S robotic arm. Two Intel D435i cameras provide input for our 3D vision policy from different perspectives, exclusively utilizing their RGB streams. One camera is mounted on the robotic arm, while the other offers a third-person view, as shown in Fig.~\ref{fig:hardware_tools} (a).


\textbf{End-Effector Tools.}
We utilize two end-effector tools: a suction cup and a parallel gripper. The suction cup, which was custom-made by us via 3D printing, is used to handle boxed objects. The parallel gripper, Flexiv GRAV model, is employed for grasping bottled objects (illustrated in Fig.~\ref{fig:hardware_tools} (b)).


\begin{figure*}[tb]
    \centering
    \includegraphics[trim=0.0cm 10cm 12.5cm 0.0cm, clip,  width=1.0\linewidth]{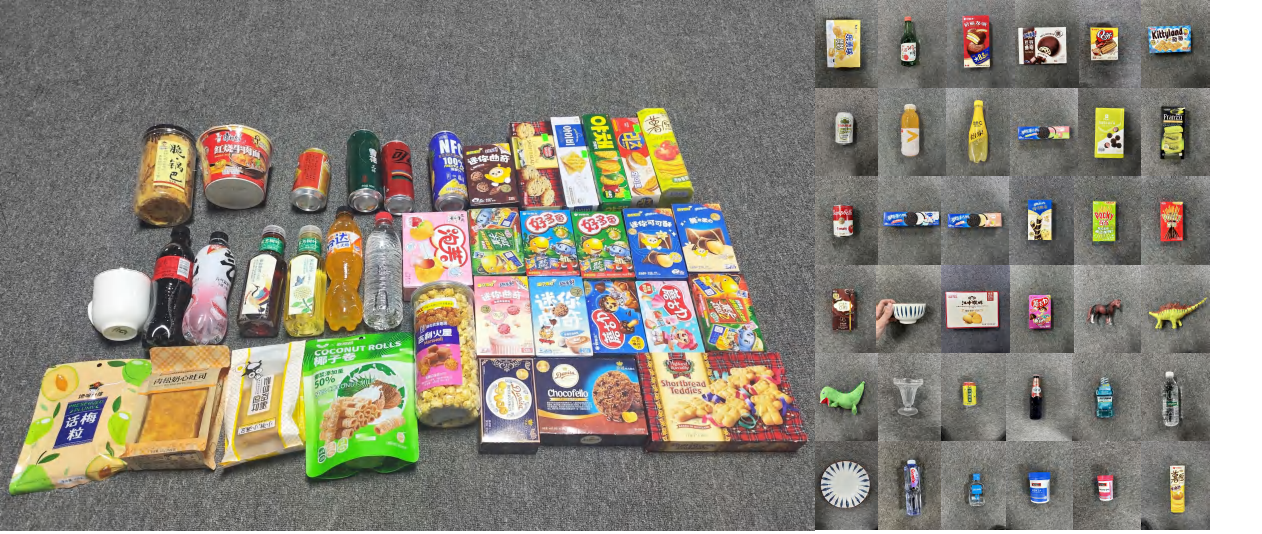}
    \caption{Objects used in real-robot experiments, including a diverse range of items with various shapes (such as boxed, bottled, and irregular forms) and materials (including transparent and reflective types).}
    \label{fig:objects}
\end{figure*}
\textbf{Experiment Scenes.}
Our main real-world experiments are primarily conducted in a replicated retail environment. Additionally, we perform experiments in settings involving a cabinet and a rack, all scenes all shown in Fig.~\ref{fig:scenes}.
And all 30 experimental layouts are shown in Fig.~\ref{fig:layouts}.

\subsection{Real-World System Design}
Our real-world system employs a \textbf{\textit{hierarchical, asynchronous control architecture}} to ensure smooth and stable robot execution. The system is decoupled into two main layers: a high-level \textbf{\textit{Policy Layer}} and a low-level \textbf{\textit{Control Layer}}.

The \textbf{\textit{Policy Layer}} operates at a low frequency of \textbf{\textit{10 Hz}}. In this layer, the policy network receives the latest sensor observations, performs inference, and generates a target end-effector (EE) pose (position and orientation). This target pose is then sent as a command to the lower-level controller.

The \textbf{\textit{Control Layer}} is an \textbf{\textit{IK-based  controller}} that runs asynchronously at a high frequency of \textbf{\textit{100 Hz}}. This layer is responsible for translating the target EE pose into a smooth trajectory of joint positions (\texttt{qpos}). It receives commands from the policy layer asynchronously and interpolates the robot's current pose towards the target. In each control cycle, it calculates a small, kinematically valid motion step using an inverse kinematics solver, ensuring that the robot's velocity limits are respected. This decoupling of low-frequency policy decisions from high-frequency motion control is crucial for stable and continuous real-world performance.

\subsection{Experiment objects}
The objects used in our experiments encompass a diverse range of types, including those that are boxed, bottled, transparent, reflective, or irregularly shaped. The experimental objects are displayed in Fig.~\ref{fig:objects}.

\begin{figure*}[tb]
    \centering
    \includegraphics[trim=0.0cm 7.5cm 6cm 0.0cm, clip,  width=1.0\linewidth]{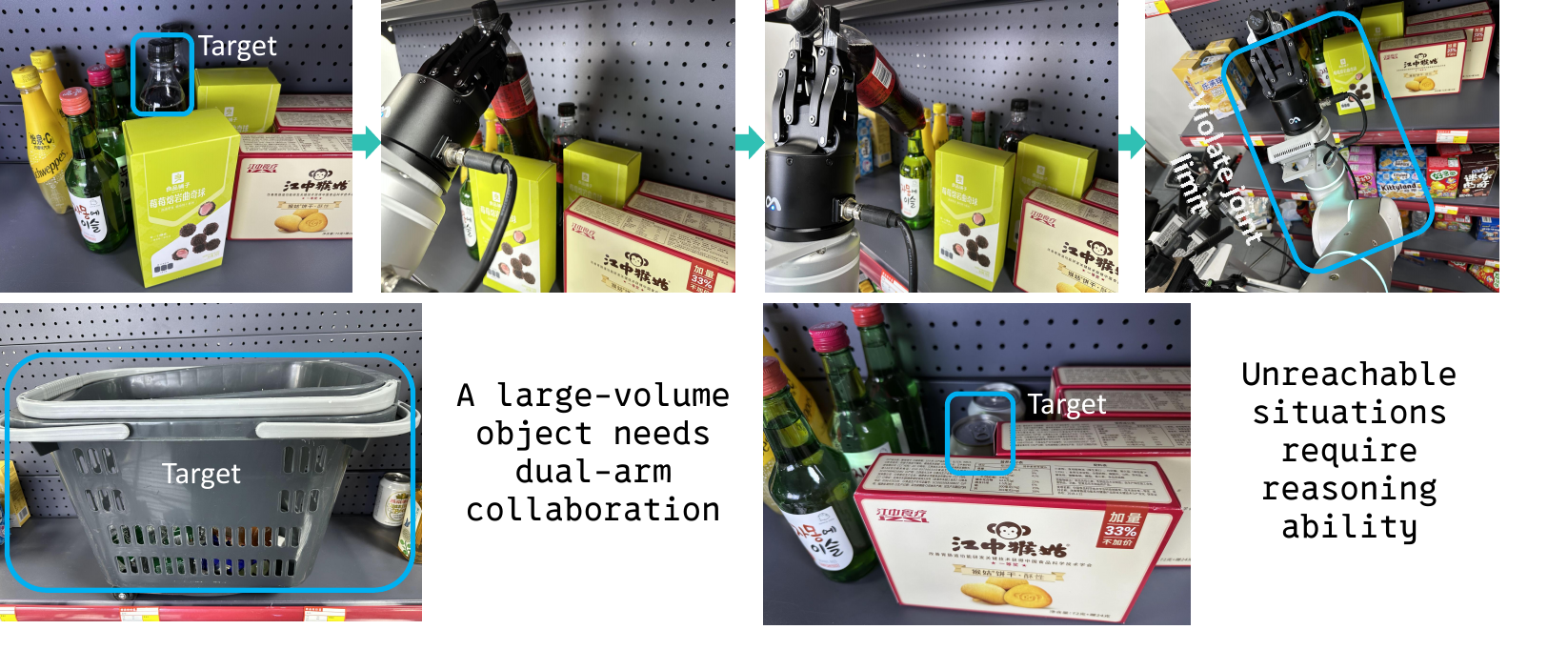}
    \caption{Illustration of Limitations. (Top row) Actions exceeding joint limits due to an attempted approach from above. (Bottom row, left) A large-volume object demonstrating the need for dual-arm manipulation. (Bottom row, right) A completely occluded and initially unreachable target, necessitating a multi-step reasoning process involving moving the obstacle, fetching the target, and subsequently returning the obstacle to its original position.}
    \label{fig:failure}
    \vspace{-1em}
\end{figure*}
\section{Failure Modes Analysis}
\label{appendix_f}
When scenes become excessively complex, there is a possibility that the policy's output will lead to joint limit violations for the robotic arm, causing the task to fail. The top of Fig.~\ref{fig:failure} depicts a scenario where the robot's intended rotational approach from above to fetch an object results in exceeding its joint limits and a collision with the upper barrier. Furthermore, when an object is too large and too heavy, we need to employ dual-arm coordination, as shown in the bottom of Fig.~\ref{fig:failure}. Finally, when the target is completely occluded to the point of being unreachable, reasoning capabilities are required to first move the obstacles aside, then retrieve the target, and subsequently restore the obstacles to their original positions (Fig.~\ref{fig:failure}). All of these can be considered as future work.

\clearpage


\end{bibunit}
\end{document}